\documentclass[10pt,twocolumn,letterpaper]{article}
\usepackage{amsmath, amssymb, amsthm}

\usepackage[table]{xcolor}    
\usepackage{subcaption}   
\usepackage{booktabs}
\usepackage{multirow}

\usepackage{fullpage}
\usepackage{graphicx}
\usepackage{multirow}

\usepackage[ruled,vlined]{algorithm2e}
\usepackage{algpseudocode}
\usepackage{mathrsfs} 
\usepackage{subcaption}
\usepackage{booktabs}

\usepackage{cvpr}              
\definecolor{cvprblue}{rgb}{0.21,0.49,0.74}
\usepackage[pagebackref,breaklinks,colorlinks,allcolors=cvprblue]{hyperref}

\def\paperID{} 
\def\confName{CVPR}
\def\confYear{2026}

\title{%
  \raisebox{-0.3\height}{\includegraphics[height=1.3em]{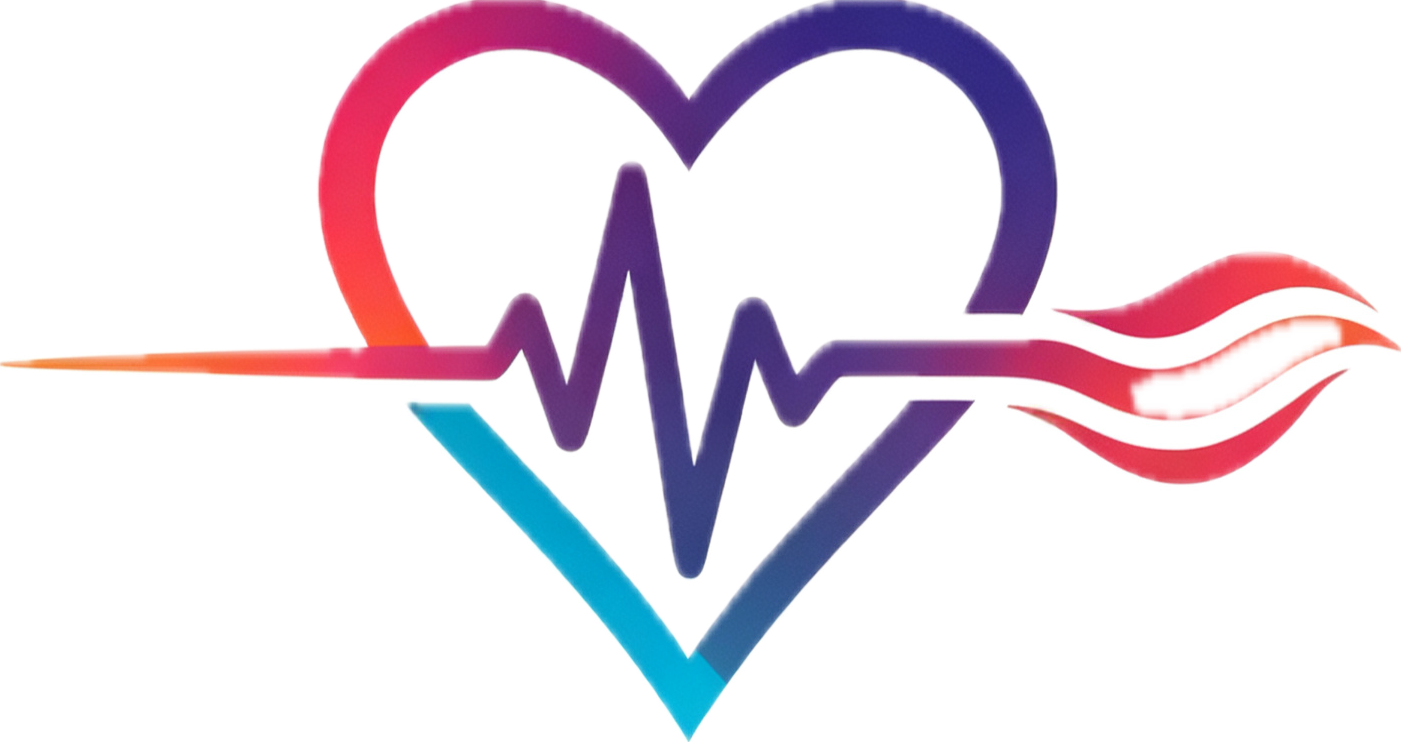}}\hspace{0.1em}%
  FLOW: Optimal Transport-Driven Feature Warping for Generalized Remote Physiological Measurement
}

\author{
    Bo Zhao$^{1*}$,  
    Junzhe Cao$^{1,5*}$, 
    Dan Guo $^{2}$,
    Dongmin Huang$^{4}$,
    Wenjin Wang $^{4}$,
    Tao Tan$^{3}$
    Yue Sun$^{3\ddagger}$, 
    Zitong Yu$^{1,6\ddagger}$ \\
    $^{1}$Great Bay University
    $^{2}$Hefei University of Technology,
    $^{3}$Macao Polytechnic University ,\\ 
    $^{4}$Southern University of Science and Technology, 
    $^{5}$Harbin Institute of Technology, Shenzhen,
    \\
    $^{6}$Dongguan Key Laboratory for Intelligence and Information Technology\\
    {\tt\small bozhao@link.cuhk.edu.cn, zitong.yu@ieee.org} \\
    $^*$ Equal contribution, $^\ddagger$ Corresponding author
}

\begin{document}
\maketitle
\begin{abstract}
Remote photoplethysmography (rPPG) enables non-contact physiological measurement but remains vulnerable to domain shifts from illumination, motion, and sensors. We propose \textbf{FLOW (Feature-Level Optimal Warping)}, an \emph{optimal transport--driven} framework for domain-generalized rPPG. FLOW integrates a \textbf{Temporal Refinement Module (TRM)} to stabilize temporal dynamics and a \textbf{Prototype-based Cross-Temporal Optimal Transport (PCOT)} module to achieve domain-invariant alignment via learnable prototypes. Beyond feature alignment, FLOW employs soft cross-temporal correspondence modeling that aligns temporal features in a flexible manner, allowing the model to respect and preserve the intrinsic rhythmic patterns of physiological signals.  Moreover, the lightweight design of our modules allows seamless integration into existing end-to-end rPPG architectures without additional preprocessing. Two regularization terms further enforce source consistency and identity preservation. Theoretically, we derive a generalization bound under conditional optimal transport. Extensive experiments across four rPPG benchmarks show that FLOW achieves state-of-the-art cross-domain performance with lightweight design and strong physiological fidelity.
\end{abstract}
    
\vspace{-0.9em}
\section{Introduction}

Remote photoplethysmography (rPPG) is a non-contact technique for estimating physiological signals such as heart rate and blood volume pulse (BVP) from facial videos. Compared with traditional contact-based photoplethysmography (PPG), rPPG offers a more convenient and hygienic alternative, making it highly attractive for applications in telemedicine, emotion recognition, and fitness monitoring. With the advent of deep learning, recent studies have demonstrated that end-to-end neural networks can directly learn complex spatial-temporal patterns from raw videos for robust rPPG signal prediction~\cite{chen2018deepphys, yu2019remote, niu2020video,cao2026physnext,wu2025cardiacmamba}.

However, despite promising results, the generalization ability of rPPG models—particularly end-to-end rPPG models—remains severely limited. These models often suffer significant performance drops when applied to new domains with different illumination, camera sensors, skin tones, or motion patterns. This issue is primarily due to domain shifts, i.e., the distributional differences between training and deployment environments. In real-world scenarios, such shifts are inevitable, and collecting labeled data from every possible target domain is impractical~\cite{huang2024etag,zhu2026H-GAR}. Hence, domain generalization (DG)~\cite{sun2016return, gretton2012kernel}, which aims to train models that generalize well to unseen domains without accessing any target data, becomes a critical challenge for making rPPG truly deployable at scale.

While domain generalization has been extensively studied in image classification and other computer vision tasks, its application to end-to-end rPPG learning remains largely unexplored. Most existing efforts in rPPG generalization either rely on data-level preprocessing like spatial-temporal map (STMap) construction followed by handcrafted pipeline tuning~\cite{wang2021domain, wang2024rppg, wang2025physmle, li2024bi}, or focus on architectural innovations within single-source settings~\cite{Du_2023_CVPR, Savic_Zhao_2024_ECCV, Xie_et_al_2024_arXiv, zeng2024survey}. These approaches, however, face two major limitations: (1) They do not address generalization for fully end-to-end pipelines, where raw video is directly mapped to physiological signals—an increasingly adopted paradigm in modern rPPG; (2) They often lack a principled theoretical understanding of how to align or unify diverse source domain representations.

To fill this gap, we propose \textbf{FLOW (Feature-Level Optimal Warping)}, a novel Optimal Transport–driven framework for \emph{multi-source domain generalization} in end-to-end rPPG learning. 
Our key insight is to interpret inter-domain variations as structured transport problems and leverage the geometry of Optimal Transport (OT)~\cite{peyre2019computational,courty2017optimal} to achieve principled feature alignment. 
Specifically, FLOW introduces a \emph{plug-and-play feature-level warping module} that aligns feature distributions across multiple source domains in a shared latent space. Unlike adversarial or statistical alignment approaches, this OT-based formulation provides an interpretable and mathematically grounded mechanism for domain-invariant representation learning while maintaining compatibility with various rPPG backbones~\cite{yu2019remote,song2021hr}.

To enhance stability and physiological fidelity, we design \textbf{FLOW} as a unified framework that integrates temporal refinement, prototype-based alignment, and consistency regularization. A lightweight \textbf{Temporal Refinement Module (TRM)} first produces temporally coherent features by suppressing high-frequency motion artifacts and spatial entanglement. The \textbf{Prototype-based Cross-Temporal Optimal Transport (PCOT)} module then performs soft temporal alignment by establishing correspondences between features and a learnable prototype bank, yielding domain-invariant yet physiologically meaningful representations. To further stabilize learning, we introduce two regularization terms: a \textbf{source-consistency constraint} encouraging uniform prototype usage across domains and an \textbf{identity-preservation constraint} maintaining proximity between original and aligned features. Together, these components form a coherent OT-driven objective that reduces inter-domain discrepancies while preserving temporal and spectral integrity crucial for accurate physiological estimation.

From a theoretical perspective, we also derive a \textbf{multi-source generalization bound} under the conditional optimal transport discrepancy, formally linking alignment quality to prediction risk on unseen domains. This provides a principled explanation for FLOW’s robustness in cross-domain temporal regression tasks.

In summary, our main contributions are as follows:
\begin{itemize}
    \item We propose \textbf{FLOW}, a architecture-agnostic framework for domain-generalized physiological signal estimation.
    \item We introduce a \textbf{Temporal Refinement Module (TRM)} that stabilizes heterogeneous temporal patterns across subjects and domains. The module learns to capture local  temporal dependencies, suppressing motion artifacts and inconsistent rhythm dynamics. 
    \item We propose a \textbf{Prototype-based Cross-Temporal Optimal Transport (PCOT)} module to align domain-specific features with a  prototype bank.PCOT formulates temporal correspondence as an OT problem with task-aware constraints, allowing flexible alignment across domains.

\end{itemize}

\section{Related Work}
\label{sec:formatting}

\subsection{Remote Photoplethysmography (rPPG)}
rPPG measurement aims to estimate physiological signals such as heart rate and blood volume pulse (BVP) using only video recordings of human faces. Early rPPG methods mainly relied on signal processing techniques to extract pulse signals from color fluctuations in skin regions~\cite{verkruysse2008remote, poh2010non, liu2024one, shi2024adaptively}. Later, more robust pipelines emerged by introducing spatial-temporal maps (STMaps)~\cite{lu2023neuron,wang2024rppg, niu2019rhythmnet}, which convert pixel-level temporal signals into images for use with convolutional neural networks (CNNs)~\cite{niu2020video, wang2021domain}. Recently, end-to-end deep learning approaches have gained attention for their ability to directly regress physiological signals from raw videos without relying on handcrafted pre-processing. Models such as DeepPhys~\cite{chen2018deepphys}, PhysNet~\cite{yu2019remote}, and PhysFormer~\cite{yu2021physformer} learn joint spatial-temporal features from video sequences in an end-to-end manner. These methods demonstrate improved performance and robustness compared to traditional pipelines, yet their generalization across domains remains a major challenge.

\subsection{Domain Generalization (DG)}
Domain generalization (DG) aims to train models that perform well on unseen domains without access to any target-domain data during training. A wide variety of DG approaches have been proposed in the fields of image classification, segmentation, and medical imaging. These include data augmentation-based strategies~\cite{zhou2021domain}, feature alignment using adversarial learning~\cite{li2018domain,huang2021unified}, and regularization-based methods~\cite{dubey2021adaptive,zhu2025emosym}. In the rPPG literature, domain shift—caused by differences in lighting, motion, skin tone, or recording hardware—has been widely observed to degrade model performance~\cite{mcduff2018survey, zhu2025uniemo,liu2025llm}. Some recent works attempt to address this by using domain-aware training protocols or specialized network architectures~\cite{wang2021domain, yu2021physformer}. However, these studies mostly focus on single-source settings or rely heavily on STMap preprocessing, limiting their applicability to end-to-end pipelines. Moreover, they often lack theoretical grounding, making it hard to generalize their findings across tasks and architectures. To the best of our knowledge, there is currently no dedicated study focusing on domain generalization for end-to-end rPPG models, despite their increasing importance in real-world applications.

\subsection{Optimal Transport for Domain Adaptation and Generalization}
Optimal Transport (OT) has emerged as a powerful tool for comparing and aligning probability distributions. In machine learning, OT has been successfully applied to unsupervised domain adaptation, particularly for image classification tasks~\cite{courty2017joint, damodaran2018deepjdot}. The Wasserstein distance~\cite{gangbo1996geometry}, in particular, offers a meaningful and geometry-aware metric to measure distributional divergence. Some studies have extended OT to domain generalization by aligning multiple source domains in a shared feature space~\cite{montesuma2021wasserstein,liu2026aullm++, huang2026complementarity}. However, these methods are mostly limited to classification problems and are not directly applicable to time-series regression tasks like rPPG measurement. Furthermore, OT has not yet been explored in the context of rPPG, especially in plug-and-play modules for end-to-end networks. In contrast, our work is the first to apply OT-based alignment to multi-source domain generalization for end-to-end rPPG models, offering both algorithmic contributions and theoretical guarantees.

\begin{figure*}[t] 
\centering 
\includegraphics[width=\textwidth]{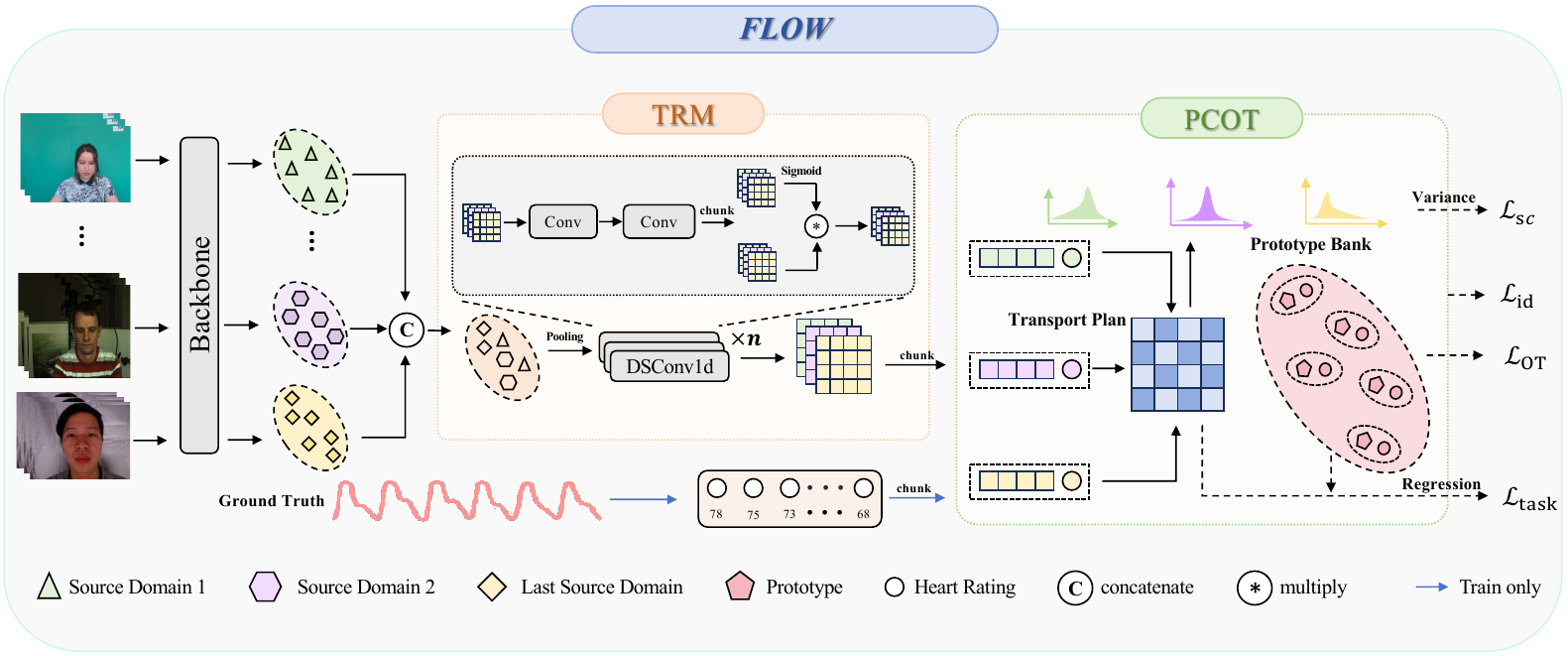} 
\vspace{-1.9em} 
\caption{
Overall architecture of FLOW. Given multi-domain video inputs, a shared backbone extracts intermediate features that are temporally refined by the TRM to suppress noise and unify dynamics. The refined sequences are then aligned through the PCOT module, which computes an optimal transport plan between temporal features and a learnable prototype bank. The framework is trained with the OT-based alignment loss $\mathcal{L}_{OT}$, the source-consistency loss $\mathcal{L}_{sc}$, the identity-preservation loss $\mathcal{L}_{id}$ and the task loss $\mathcal{L}_{task}$.
During training, the ground truth heart rate signals are used to guide and refine the transport plan for better prototype alignment. During inference, the model directly applies the learned transport plan to predict heart rates, without using any ground truth signals.
}
\label{fig:fig_2} 
\vspace{-0.8em} 
\end{figure*}

\section{Methodology}

Our goal is to learn temporally stable and physiologically coherent representations that remain robust across subjects and recording conditions. 
To achieve this, we propose a unified architecture in which temporal refinement precedes cross-domain alignment.
As illustrated in Figure~\ref{fig:fig_2}, intermediate representations are first processed by a lightweight \textbf{Temporal Refinement Module (TRM)}, which unifies heterogeneous spatiotemporal features into a consistent temporal form.
The refined features are then aligned via the \textbf{Prototype-based Cross-Temporal Optimal Transport (PCOT)} module, which establishes soft correspondences between temporal features and a shared prototype bank.
Finally, a set of regularization objectives ensures temporal consistency, cross-domain coherence, and physiological interpretability.

\subsection{Prototype-based Cross-Temporal Optimal Transport (PCOT)}

Temporal signals from different domains often exhibit heterogeneous rhythms and domain-specific distortions. 
To achieve domain-invariant alignment, PCOT models each temporal step as a distribution over shared prototypes and aligns these representations via an \textbf{entropic optimal transport (OT)} formulation. 
This yields soft correspondences between time-varying features and domain-agnostic prototypes.

\vspace{0.3em}
\noindent\textbf{Prototype construction.} \quad
We maintain a learnable set of prototypes $\mathcal{P} = \{p_k\}_{k=1}^K$ and their associated physiological anchors $\mathcal{H} = \{h_k\}_{k=1}^K$. 
Given a batch of temporal features $X = \{x_t\}_{t=1}^T$ ($x_t \in \mathbb{R}^C$), each temporal step is softly matched to prototypes based on both feature similarity and physiological coherence. 
The transport cost is defined as:
\begin{equation}
C_{t,k} = \|W(x_t - p_k)\|_2^2 + 
\lambda_{\mathrm{hr}}\!\left(1 - \exp\!\left[-\frac{(h_t - h_k)^2}{2\sigma^2}\right]\right),
\end{equation}
where $W$ is a learnable diagonal weighting matrix, and $h_t$ is the estimated heart rate from an auxiliary head $\mathrm{Head}_{\text{HR}}$. 
The first term enforces semantic similarity, while the second term penalizes physiological inconsistency, encouraging prototypes to encode \textbf{domain-invariant but physiologically coherent} features.

\vspace{0.4em}
\noindent\textbf{Optimal transport formulation.} \quad
Let $\mu$ and $\nu$ denote the empirical distributions of temporal features and prototypes, where $\mu_t = 1/T$ and $\nu_k = 1/K$. 
The matching is formalized as an entropic OT problem:
\begin{equation}
S_\varepsilon(\mu,\nu)
= \min_{\Pi \in \mathcal{U}(\mu,\nu)}
\langle C, \Pi \rangle + \varepsilon\, H(\Pi),
\end{equation}
where $\mathcal{U}(\mu,\nu) = \{\Pi \ge 0 \mid \Pi\mathbf{1}=\mu,\, \Pi^\top\mathbf{1}=\nu\}$ 
and $H(\Pi) = -\sum_{t,k}\pi_{t,k}(\log \pi_{t,k}-1)$ denotes the entropy regularizer. 
The regularization coefficient $\varepsilon$ controls smoothness and ensures differentiability.

\vspace{0.4em}
\noindent\textbf{Sinkhorn normalization.} \quad
We compute the optimal coupling $\Pi^\star$ using the Sinkhorn algorithm~\cite{cuturi2013sinkhorn}:
\begin{equation}
K = \exp\!\left(-\frac{C}{\varepsilon}\right), \quad
\Pi^\star = \mathrm{Diag}(u)\, K\, \mathrm{Diag}(v),
\end{equation}
with iterative updates:
\begin{equation}
u^{(m+1)} = \frac{a}{K v^{(m)}}, \quad
v^{(m+1)} = \frac{b}{K^{\!\top}u^{(m+1)}},
\end{equation}
where $a_t=\tfrac{1}{T}$ and $b_k=\tfrac{1}{K}$. 
This normalization ensures marginal consistency ($\Pi^\star\mathbf{1}=a$, $(\Pi^\star)^\top\mathbf{1}=b$) and provides a stable and differentiable transport plan.

\vspace{0.4em}
\noindent\textbf{Barycentric projection and alignment.} \quad
The optimal coupling $\Pi^\star$ defines soft correspondences between temporal steps and prototypes. 
Aligned representations are obtained by barycentric projection:
\begin{equation}
\tilde{x}_t = \sum_{k=1}^K \pi^\star_{t,k} p_k, \quad
\tilde{X} = \{\tilde{x}_t\}_{t=1}^T.
\end{equation}
This projection re-expresses temporal features in the prototype manifold, removing domain-specific variations and improving temporal smoothness. 
To mitigate entropy bias, we adopt the \textbf{debiased Sinkhorn divergence} as the alignment loss:
\begin{equation}
\mathcal{L}_{\text{OT}} = 
S_\varepsilon(\mu,\nu)
- \tfrac{1}{2} S_\varepsilon(\mu,\mu)
- \tfrac{1}{2} S_\varepsilon(\nu,\nu).
\end{equation}
This symmetric and unbiased discrepancy measure stabilizes alignment and enhances generalization.

Intuitively, PCOT constructs a compact and domain-agnostic prototype manifold, where temporal features are softly aligned via differentiable OT. 
The resulting mapping disentangles domain-specific appearance factors while preserving intrinsic physiological rhythms.

\subsection{Temporal Refinement Module (TRM)}

While PCOT aligns temporal dynamics across domains, intermediate features may still contain spatial entanglement or high-frequency noise. 
The \textbf{Temporal Refinement Module (TRM)} addresses this issue by converting heterogeneous feature tensors into a unified temporal form and refining local dynamics through lightweight depthwise-separable convolutions.

\vspace{0.4em}
\noindent\textbf{Unified temporal representation.} \quad
Given intermediate features $F$ of arbitrary shape (e.g., $B\times C\times H\times W$ or $B\times C\times T\times H\times W$), TRM performs spatial global pooling to produce a consistent temporal sequence:
\begin{equation}
X = \mathrm{Pool}_{\text{spatial}}(F) \in \mathbb{R}^{B\times T\times C}.
\end{equation}
This ensures all temporal tokens share a unified semantic basis for subsequent modeling.

\vspace{0.4em}
\noindent\textbf{Local temporal modeling.} \quad
TRM refines temporal sequences using stacked depthwise-separable 1D convolutional blocks:
\begin{equation}
Y^{(l+1)} = \mathrm{Norm}\!\left(X^{(l)} + \mathcal{F}_{\text{TRM}}\!\left(X^{(l)}\right)\right),
\quad Y^{(0)} = X.
\end{equation}
Each block comprises:

\textbf{(1) Depthwise temporal filtering:}
\begin{equation}
Z_{t,c} = \sum_{i=1}^{k} w^{(d)}_{c,i}\, X^{(l)}_{t+i,c},
\end{equation}
which captures localized rhythmic dependencies.

\textbf{(2) Pointwise channel fusion:}
\begin{equation}
\tilde{Z}_t = \phi(W_p\, Z_t + b_p),
\end{equation}
where $W_p \in \mathbb{R}^{C\times C}$ and $\phi(\cdot)$ denotes a nonlinearity (e.g., GELU).  
The combined transformation is:
\begin{equation}
\mathcal{F}_{\text{TRM}}(X^{(l)}) = \phi\!\big(W_p * (W_d * X^{(l)})\big),
\end{equation}
balancing local temporal filtering and inter-channel fusion with complexity $\mathcal{O}(BTCk)$. 
Stacking multiple layers progressively suppresses short-term noise and refines rhythmic patterns:
\begin{equation}
X^{\text{TRM}} = \mathrm{TRM}(X) \in \mathbb{R}^{B\times T\times C}.
\end{equation}
From a signal-processing perspective, TRM acts as a low-pass temporal filter that enhances phase stability and rhythmic coherence.

\begin{table*}[t]
\vspace{-1.8em}
\centering
\caption{Multi-domain generalization evaluation. Best results are marked in \textbf{bold}. `+' means domain generalization methods are based on `Baseline'.}
\vspace{-0.5em}
\label{tab:cross_dataset}

\resizebox{\textwidth}{!}{
\begin{tabular}{lccccccccccccccc}
\toprule
\textbf{Model} & \multicolumn{3}{c}{\textbf{Others$\rightarrow$U}} & \multicolumn{3}{c}{\textbf{Others$\rightarrow$P}} & \multicolumn{3}{c}{\textbf{Others$\rightarrow$B}} & \multicolumn{3}{c}{\textbf{Others$\rightarrow$M}} & \multicolumn{3}{c}{\textbf{Average}} \\
 \cmidrule(lr){2-4} \cmidrule(lr){5-7}  \cmidrule(lr){8-10} \cmidrule(lr){11-13} \cmidrule(lr){14-16}
& MAE$\downarrow$ & RMSE$\downarrow$ & R$\uparrow$
& MAE$\downarrow$ & RMSE$\downarrow$ & R$\uparrow$
& MAE$\downarrow$ & RMSE$\downarrow$ & R$\uparrow$
& MAE$\downarrow$ & RMSE$\downarrow$ & R$\uparrow$
& \textbf{MAE$\downarrow$} & \textbf{RMSE$\downarrow$} & \textbf{R$\uparrow$} \\
\midrule
Green~\cite{verkruysse2008remote}  & 19.73 & 31.00 & 0.37 & 10.09 & 23.85 & 0.34 & 6.89 & 10.39 & 0.60 & 21.68 & 27.69 & -0.01 & 14.10 & 23.73 & 0.33 \\
CHROM~\cite{de2013robust}  & 7.23 & 8.92 & 0.51 & 9.79 & 12.76 & 0.37 & 6.09 & 8.29 & 0.51 & 13.66 &18.76 &  0.08 & 9.69 & 12.68 & 0.37 \\
POS~\cite{wang2016algorithmic} & 7.35 & 8.04 & 0.49 & 9.82 & 13.44 & 0.34 & 5.04 & 7.12 & 0.63 & 12.36  &17.71 &  0.18 & 8.64 & 11.58 & 0.41 \\
\midrule
EfficientPhys~\cite{liu2023efficientphys} & 12.87 & 18.80 & 0.19 & 7.15 & 15.04 & 0.23 & 32.30 & 34.00 & -0.03 & 12.87 & 18.80 & 0.19 & 16.80 & 21.66 & 0.14 \\
PhysFormer~\cite{yu2022physformer} & 10.29 & 18.13 & 0.60 & 19.75 & 24.30 & 0.24 & 22.09 & 26.21 & 0.03 & 13.90 & 19.30 & 0.06 & 16.51 & 21.98 & 0.23 \\
PhysNet~\cite{yu2019remote}  & 13.83 & 23.66 & 0.35 & 33.23 & 35.25 & -0.15 & 12.75 & 16.37 & 0.08 & 13.37 & 16.64 & 0.29 & 18.30 & 22.98 & 0.14 \\
RhythmFormer~\cite{zou2025rhythmformer}  & 14.71 & 22.49 & 0.43 & 21.11 & 25.76 & 0.04 & 6.04 & 10.84 & 0.42 & 16.14 & 20.50 & -0.11 & 14.50 & 19.90 & 0.20 \\
\midrule
NEST~\cite{lu2023neuron}  & 12.24 & 10.56 & 0.36 & 19.26 & 26.11 & -0.06 & 9.19 & 12.38 & 0.19 & 13.97 & 18.20 & 0.15 & 13.67 & 16.81 & 0.16 \\
Greip~\cite{zhang2025advancing}  & 17.50 & 20.42 & 0.21 & 5.07 & 14.50 & 0.78 & 7.94 & 10.93 & -0.03 & 13.02 & 17.11 & 0.14 & 10.88 & 15.74 & 0.28 \\
\midrule
Baseline~\cite{xie2025physllm}  & 9.92 & 13.92 & 0.64 & 15.97 & 26.61 & 0.23 & 6.02 & 8.61 & 0.63 & 12.23 & 15.51 & 0.15 & 11.04 & 16.16 & 0.41 \\
Coral+~\cite{sun2016return} & 11.47 & 14.42 & 0.54 & 14.18 & 23.04 & 0.26 & 3.06 & 4.42 & 0.95 & 9.18 & 14.96 & 0.43 & 9.97 & 14.21 & 0.55 \\
MMD+~\cite{gretton2012kernel} & 9.18 & 12.14 & 0.60 & 16.56 & 22.93 & 0.28 & 2.80 & 3.67 & 0.95 & 8.87 & 14.39 & 0.45 & 9.35 & 13.28 & 0.57 \\
\rowcolor{blue!10}
\textbf{FLOW (Ours)} & \textbf{6.89} & \textbf{10.12} & \textbf{0.69} & \textbf{10.86} & \textbf{16.47} & \textbf{0.64} & \textbf{2.23} & \textbf{3.36} & \textbf{0.97} & \textbf{7.38} & \textbf{13.12} & \textbf{0.51} & \textbf{6.84} & \textbf{10.75} & \textbf{0.70} \\
\bottomrule
\end{tabular}
}
\end{table*}

\subsection{Regularization for Stable Cross-Domain Alignment}

Although OT provides a principled alignment mechanism, large domain gaps can lead to unstable or over-smoothed transport plans. 
We introduce two complementary regularization terms to enhance robustness and preserve physiological interpretability.

\vspace{0.4em}
\noindent\textbf{Source-consistency regularization.} \quad
For each source domain $D_j$, we compute the mean prototype-assignment histogram based on the optimal transport plan $\Pi^\star \in \mathbb{R}^{T \times K}$:
\begin{equation}
\bar{\mathbf{h}}_j = \frac{1}{|D_j|} \sum_{i \in D_j} 
\frac{1}{T} \sum_{t=1}^{T} \pi^{\star}_{t,k}.
\end{equation}
To ensure consistent prototype utilization across domains, we minimize the variance among these mean histograms:
\begin{equation}
\mathcal{L}_{\text{src}} = 
\frac{1}{M} \sum_{j=1}^{M} 
\| \bar{\mathbf{h}}_j - \bar{\mathbf{h}} \|_2^2,
\quad
\bar{\mathbf{h}} = \frac{1}{M} \sum_{j=1}^{M} \bar{\mathbf{h}}_j.
\end{equation}
This encourages all domains to share a uniform prototype occupancy pattern, reinforcing domain-invariant semantics.

\vspace{0.4em}
\noindent\textbf{Identity-preservation regularization.} \quad
To prevent excessive feature deformation, we constrain the distance between original and aligned representations:
\begin{equation}
\mathcal{L}_{\text{id}} = 
\frac{1}{BT} \sum_{b=1}^{B}\sum_{t=1}^{T}
\|\tilde{x}_{b,t} - x_{b,t}\|_2^2.
\end{equation}
This penalty stabilizes alignment while preserving subject identity and intrinsic rhythm.

\vspace{0.2em}
\noindent\textbf{Final objective.} \quad
The overall training objective integrates OT alignment with the two regularization terms:
\begin{equation}
\mathcal{L}_{\text{total}} =\mathcal{L}_{\text{task}}+
\lambda_{\text{OT}}\mathcal{L}_{\text{OT}} +
\lambda_{\text{src}} \mathcal{L}_{\text{src}} +
\lambda_{\text{id}} \mathcal{L}_{\text{id}}.
\end{equation}
Here, $\lambda_{\text{src}}$ and $\lambda_{\text{id}}$ balance alignment flexibility and representation stability. The detailed model training and inference process can be found in the Appendix ~\ref{train}.

Together, TRM, PCOT, and the regularization objectives form a coherent pipeline: 
TRM unifies and refines temporal dynamics, PCOT aligns features through prototype-guided optimal transport, 
and the regularizations ensure both global domain consistency and local physiological preservation. 
This design enables smooth temporal representations without adversarial training.

\section{Experiments}
We conduct multi-source domain generalization experiments on four public datasets: UBFC-rPPG (U)~\cite{Bobbia2017UBFCrPPG}, PURE (P)~\cite{Stricker2014PURE}, BUAA-MIHR (B)~\cite{Xi2020BUAAMIHR}, and MMPD (M)~\cite{tang2023mmpd}. We report Mean Absolute Error (MAE), Root Mean Square Error (RMSE), and Pearson correlation coefficient (R), where lower MAE/RMSE and higher R indicate better performance. All experimental settings and implementation details are included in Appendix~\ref{settings}.

\begin{table*}[!t]
\centering
\caption{Limited-source domain generalization performance on MMPD. Best results are in \textbf{bold}.}
\vspace{-0.5em}
\label{tab:limited_mmpd}
\renewcommand\arraystretch{1}
\resizebox{\textwidth}{!}{
\begin{tabular}{lcccccccccccc}
\toprule
\textbf{Model} &
\multicolumn{3}{c}{\textbf{P+B}} &
\multicolumn{3}{c}{\textbf{P+U}} &
\multicolumn{3}{c}{\textbf{B+U}} &
\multicolumn{3}{c}{\textbf{Average}} \\
\cmidrule(lr){2-4}\cmidrule(lr){5-7}\cmidrule(lr){8-10}\cmidrule(lr){11-13}
& MAE$\downarrow$ & RMSE$\downarrow$ & R$\uparrow$
& MAE$\downarrow$ & RMSE$\downarrow$ & R$\uparrow$
& MAE$\downarrow$ & RMSE$\downarrow$ & R$\uparrow$
& MAE$\downarrow$ & RMSE$\downarrow$ & R$\uparrow$ \\
\midrule
PhysNet~\cite{yu2019remote} & 13.20 & 16.70 & 0.23 & 11.00 & 17.30 & 0.28 & 13.50 & 17.00 & 0.09 & 12.57 & 17.00 & 0.20 \\
PhysFormer~\cite{yu2022physformer} & 13.90 & 18.60 & 0.21 & 11.40 & 17.50 & 0.23 & 13.20 & 16.50 & 0.12 & 12.83 & 17.53 & 0.19 \\
EfficientPhys~\cite{liu2023efficientphys} & 11.90 & 18.50 & 0.21 & 11.80 & 18.90 & 0.22 & 15.50 & 20.80 & 0.03 & 13.07 & 19.40 & 0.15 \\
RhythmFormer~\cite{zou2025rhythmformer} & 13.98 & 19.46 & 0.12 & 10.50 & 16.72 & 0.28 & 11.70 & 16.56 & 0.18 & 12.06 & 17.58 & 0.19 \\
\midrule
NEST~\cite{liu2023rppg} & 10.85 & 15.45 & 0.33 & 9.90 & 14.82 & 0.36 & 10.64 & 15.12 & 0.31 & 10.46 & 15.13 & 0.33 \\
Greip~\cite{zhang2025advancing} & 10.60 & 15.12 & 0.35 & 9.78 & 14.63 & 0.38 & 10.42 & 14.85 & 0.33 & 10.27 & 14.87 & 0.35 \\
\midrule
Baseline~\cite{xie2025physllm} & 11.90 & 15.30 & 0.26 & 9.95 & 14.96 & 0.31 & 12.10 & 15.20 & 0.21 & 11.32 & 15.15 & 0.26 \\
Coral+~\cite{sun2016return} & 10.88 & 15.47 & 0.29 & 11.62 & 15.89 & 0.25 & 10.95 & 15.34 & 0.28 & 11.15 & 15.57 & 0.27 \\
MMD+~\cite{gretton2012kernel} & 9.96 & 15.01 & 0.34 & 11.49 & 16.02 & 0.24 & 10.78 & 15.42 & 0.27 & 10.74 & 15.48 & 0.28 \\
\rowcolor{blue!10}
\textbf{FLOW (Ours)} & \textbf{8.90} & \textbf{13.64} & \textbf{0.47} & \textbf{8.34} & \textbf{12.79} & \textbf{0.51} & \textbf{8.70} & \textbf{13.36} & \textbf{0.45} & \textbf{8.65} & \textbf{13.26} & \textbf{0.48} \\
\bottomrule
\end{tabular}
}
\vspace{-0.6em}
\end{table*}

\begin{table*}[!t]
\centering
\caption{Limited-source domain generalization performance on BUAA-MIHR. Best results are in \textbf{bold}.}
\vspace{-0.5em}
\label{tab:limited_buaa}
\renewcommand\arraystretch{1}
\resizebox{\textwidth}{!}{
\begin{tabular}{lcccccccccccc}
\toprule
\textbf{Model} &
\multicolumn{3}{c}{\textbf{P+M}} &
\multicolumn{3}{c}{\textbf{M+U}} &
\multicolumn{3}{c}{\textbf{P+U}} &
\multicolumn{3}{c}{\textbf{Average}} \\
\cmidrule(lr){2-4}\cmidrule(lr){5-7}\cmidrule(lr){8-10}\cmidrule(lr){11-13}
& MAE$\downarrow$ & RMSE$\downarrow$ & R$\uparrow$
& MAE$\downarrow$ & RMSE$\downarrow$ & R$\uparrow$
& MAE$\downarrow$ & RMSE$\downarrow$ & R$\uparrow$
& MAE$\downarrow$ & RMSE$\downarrow$ & R$\uparrow$ \\
\midrule
PhysNet~\cite{yu2019remote} & 20.97 & 24.75 & 0.01 & 11.40 & 16.72 & 0.14 & 15.34 & 21.48 & -0.29 & 15.90 & 20.98 & -0.05 \\
PhysFormer~\cite{yu2022physformer} & 14.86 & 18.26 & 0.03 & 10.87 & 16.20 & 0.08 & 18.23 & 22.17 & 0.07 & 14.65 & 18.88 & 0.06 \\
EfficientPhys~\cite{liu2023efficientphys} & 4.15 & 7.14 & 0.77 & 3.00 & 5.18 & 0.89 & 3.00 & 5.18 & 0.89 & 3.38 & 5.83 & 0.85 \\
RhythmFormer~\cite{zou2025rhythmformer} & 3.55 & 5.35 & 0.90 & 6.20 & 11.23 & 0.49 & 3.90 & 6.51 & 0.82 & 4.55 & 7.70 & 0.74 \\
\midrule
NEST~\cite{liu2023rppg} & 3.32 & 5.04 & 0.92 & 3.71 & 5.67 & 0.87 & 3.15 & 4.88 & 0.93 & 3.39 & 5.20 & 0.91 \\
Greip~\cite{zhang2025advancing} & 3.21 & 4.92 & 0.93 & 3.82 & 5.86 & 0.85 & 3.08 & 4.73 & 0.94 & 3.37 & 5.17 & 0.91 \\
\midrule
Baseline~\cite{xie2025physllm} & 3.06 & 3.99 & 0.95 & 4.46 & 8.85 & 0.60 & 3.64 & 8.04 & 0.71 & 3.72 & 6.96 & 0.75 \\
Coral+~\cite{sun2016return} & 3.20 & 4.30 & 0.92 & 4.80 & 9.00 & 0.58 & 3.54 & 6.12 & 0.78 & 3.85 & 6.47 & 0.76 \\
MMD+~\cite{gretton2012kernel} & 2.98 & 3.92 & 0.95 & 4.83 & 9.15 & 0.57 & 3.89 & 6.72 & 0.72 & 3.90 & 6.60 & 0.75 \\
\rowcolor{blue!10}
\textbf{FLOW (Ours)} & \textbf{2.56} & \textbf{3.35} & \textbf{0.98} & \textbf{2.78} & \textbf{2.96} & \textbf{0.95} & \textbf{2.33} & \textbf{3.10} & \textbf{0.97} & \textbf{2.56} & \textbf{3.14} & \textbf{0.97} \\
\bottomrule
\end{tabular}
}
\vspace{-0.6em}
\end{table*}

\subsection{Experimental Results}

\subsubsection{Multi-source Domain Generalization Results}
\textbf{Comparison with standard baselines.}
We first compare FLOW with conventional signal-processing and learning-based approaches that lack explicit domain generalization mechanisms. Traditional handcrafted methods---Green~\cite{gretton2012kernel}, CHROM~\cite{de2013robust}, and POS~\cite{wang2016algorithmic}---show strong domain sensitivity, exhibiting large performance fluctuations across datasets. For instance, although CHROM achieves a reasonable MAE of 6.09 on BUAA-MIHR~\cite{Xi2020BUAAMIHR}, its error sharply increases to 13.66 on MMPD~\cite{tang2023mmpd} and 7.23 on UBFC-rPPG~\cite{Bobbia2017UBFCrPPG}, highlighting its vulnerability to illumination and motion changes.

Among deep-learning-based baselines, PhysNet~\cite{yu2019remote} and EfficientPhys~\cite{liu2023efficientphys} also exhibit poor cross-domain robustness, particularly when the target distribution deviates from training sources. Notably, PhysNet even produces a negative correlation on PURE~\cite{Stricker2014PURE} (R = -0.15), indicating a failure to capture meaningful physiological dynamics. PhysFormer~\cite{yu2022physformer} demonstrates partial robustness but still performs inconsistently across domains (e.g., R = 0.03 on BUAA-MIHR and R = 0.06 on MMPD). Overall, these results confirm that, without explicit domain-invariant modeling, existing rPPG frameworks struggle to maintain stability under distribution shifts.

\vspace{0.3em}
\noindent\textbf{Comparison with DG baselines.} \quad
We further evaluate FLOW against representative domain generalization (DG) methods---CORAL~\cite{sun2016return} and MMD~\cite{gretton2012kernel}---implemented on the same backbone for a fair comparison. Coral~\cite{sun2016return} improves correlation on PURE~\cite{Stricker2014PURE} and BUAA-MIHR~\cite{Xi2020BUAAMIHR} (e.g., R = 0.64 and R = 0.95, respectively), yet its MAE and RMSE remain relatively high, suggesting that second-order feature alignment alone is insufficient for robust temporal generalization. MMD~\cite{gretton2012kernel} yields more balanced performance across metrics but still suffers from instability on UBFC-rPPG and MMPD. In contrast, FLOW achieves consistently superior results across all unseen domains. On BUAA-MIHR, FLOW attains an MAE of 2.23 and R of 0.97, surpassing the best DG baseline (MMD~\cite{gretton2012kernel}, MAE = 2.80, R = 0.95). On UBFC-rPPG~\cite{Bobbia2017UBFCrPPG} and PURE~\cite{Stricker2014PURE}, FLOW improves the Pearson correlation by more than 0.4 compared with MMD~\cite{gretton2012kernel}, demonstrating strong robustness against appearance and motion variability.

These results highlight that effective domain generalization in rPPG requires not only statistical distribution alignment but also semantic and physiological consistency---objectives explicitly enforced in FLOW through its prototype-based optimal-transport alignment and temporal refinement modules.

\subsubsection{Limited-source Domain Generalization Results} 
To further evaluate the robustness of our method under limited training data scenarios, we perform cross-domain experiments where only two datasets were used as source domains. We selected two target datasets, MMPD~\cite{tang2023mmpd} and BUAA-MIHR~\cite{Xi2020BUAAMIHR}, due to their challenging domain characteristics. Table~\ref{tab:limited_mmpd} and Table~\ref{tab:limited_buaa} reports detailed comparisons.

\vspace{0.3em}
\noindent\textbf{Performance on MMPD~\cite{tang2023mmpd}.} \quad
Across all source-domain combinations, our method consistently achieves the lowest MAE and RMSE. When trained on PURE~\cite{Stricker2014PURE}+UBFC-rPPG~\cite{Bobbia2017UBFCrPPG}, our model attains an MAE of 9.06 and an RMSE of 14.23, outperforming both traditional approaches (e.g., Green~\cite{verkruysse2008remote}) and domain generalization baselines (e.g., Coral~\cite{sun2016return}, MMD~\cite{gretton2012kernel}). This demonstrates the model’s capability to generalize to domains with complex motion patterns and severe illumination variations, even when the diversity of training data is limited.

\vspace{0.3em}
\noindent\textbf{Performance on BUAA-MIHR~\cite{Xi2020BUAAMIHR}.} \quad
Our approach also exhibits strong generalization to BUAA-MIHR~\cite{Xi2020BUAAMIHR}. Particularly under the PURE~\cite{Stricker2014PURE}+MMPD~\cite{tang2023mmpd} training setup, it achieves an MAE of 2.56 and R = 0.98—the best among all methods. Such results highlight our framework’s ability to learn semantically consistent, domain-invariant temporal features from disjoint source domains, effectively bridging large inter-domain gaps.

\vspace{0.3em}
\noindent\textbf{Comparison with DG baselines.} \quad
Although Coral~\cite{sun2016return} and MMD~\cite{gretton2012kernel} improve over the naive baseline, their gains are often inconsistent and sensitive to domain combinations. For instance, Coral~\cite{sun2016return} performs relatively well on BUAA-MIHR~\cite{Xi2020BUAAMIHR} but degrades noticeably on MMPD~\cite{tang2023mmpd}, while MMD~\cite{gretton2012kernel} struggles when the domain discrepancy is large (e.g., PURE~\cite{Stricker2014PURE}+BUAA-MIHR~\cite{Xi2020BUAAMIHR} $\rightarrow$ MMPD~\cite{tang2023mmpd}). In contrast, our method maintains stable performance across all configurations, indicating better scalability and robustness under data-constrained scenarios.

Overall, these findings confirm that our model generalizes effectively to unseen domains even when the source diversity is limited. We attribute this robustness to our prototype-based feature alignment and temporal refinement, which jointly preserve semantic structures while reducing domain-specific bias. Consequently, our framework provides a more reliable foundation for real-world cross-domain rPPG deployment.

\subsubsection{Generalization of FLOW}
\label{sec:flow_generalization}

To assess the generalization ability of FLOW beyond a specific architecture,  we integrate it into multiple representative rPPG backbones, including RhythmFormer~\cite{zou2025rhythmformer}, EfficientPhys~\cite{liu2023efficientphys}, PhysFormer~\cite{yu2021physformer}, and PhysNet~\cite{yu2019remote}. 
As shown in Figure~\ref{fig:fig1}, FLOW consistently improves performance across diverse architectures and datasets , reducing MAE by a large margin. 
These results demonstrate that FLOW is architecture-agnostic and can be seamlessly integrated into various temporal models 
to enhance temporal stability and physiological consistency.

Furthermore, we compare FLOW with conventional domain-alignment techniques—MMD~\cite{gretton2012kernel} and Coral~\cite{sun2016return}—under the same PhysFormer backbone. 
As presented in Table~\ref{tab:flow_ablation_compact}, FLOW achieves the lowest MAE/RMSE across multiple source–target settings 
(e.g., PM-B, MU-B, and BU-M). 
While MMD~\cite{gretton2012kernel} and Coral~\cite{sun2016return} only perform statistical alignment on global feature distributions, 
FLOW introduces temporal refinement and prototype-based optimal transport, enabling more stable and physiologically meaningful cross-domain adaptation.


\begin{table}[t]
\centering
\caption{
Comparison of different domain-alignment modules applied to the PhysFormer backbone (values are MAE$\downarrow$/RMSE$\downarrow$). 
“MMD” and “CORAL” are standard alignment baselines, and “FLOW” is our proposed method.
}
\vspace{-0.5em}

\label{tab:flow_ablation_compact}

\small
\setlength{\tabcolsep}{3pt}

\begin{subtable}{\columnwidth}
\centering
\caption{Limited-source domain generalization performance on BUAA-MIHR}
\resizebox{\columnwidth}{!}{%
\begin{tabular}{lccc}
\toprule
& \textbf{P+M} & \textbf{U+M} & \textbf{P+U} \\
\midrule
PhysFormer               & 14.86/18.26 & 10.87/16.20 &  8.23/22.17 \\
PhysFormer (MMD)         & 33.12/35.26 &  8.14/11.23 & 18.41/21.10 \\
PhysFormer (CORAL)       & 13.25/20.03 & 10.69/15.24 & 14.35/19.31 \\
\rowcolor{blue!10}
\textbf{PhysFormer (FLOW)} & \textbf{10.38/14.14} & \textbf{7.18/11.14} & \textbf{9.59/13.85} \\
\bottomrule
\end{tabular}%
}
\end{subtable}


\begin{subtable}{\columnwidth}
\centering
\caption{Limited-source domain generalization performance on MMPD}
\resizebox{\columnwidth}{!}{%
\begin{tabular}{lccc}
\toprule
& \textbf{P+B} & \textbf{P+U} & \textbf{B+U} \\
\midrule
PhysFormer               & 13.91/18.63 & 11.41/17.52 & 13.20/16.50 \\
PhysFormer (MMD)         & 15.31/19.01 & 22.12/26.41 & 13.26/16.53 \\
PhysFormer (CORAL)       & 13.24/18.17 & 10.31/16.20 & 12.76/18.14 \\
\rowcolor{blue!10}
\textbf{PhysFormer (FLOW)} & \textbf{10.34/17.05} & \textbf{9.18/15.50}  & \textbf{10.01/13.72} \\
\bottomrule
\end{tabular}%
}
\end{subtable}
\vspace{-1.3em}
\end{table}

\noindent\textbf{Performance across source combinations.} \quad
In all configurations, incorporating FLOW consistently improves over the baseline PhysFormer and its DG variants. For instance, under the PURE~\cite{Stricker2014PURE}+MMPD~\cite{tang2023mmpd} training setup, FLOW reduces MAE from 14.86 to 10.38 and RMSE from 18.26 to 14.14, achieving a clear performance gain. Similarly, in the PURE~\cite{Stricker2014PURE}+BUAA-MIHR~\cite{Xi2020BUAAMIHR} setting (targeting MMPD~\cite{tang2023mmpd}), FLOW improves the correlation from 0.21 to 0.42. In contrast, both MMD~\cite{gretton2012kernel} and Coral~\cite{sun2016return} exhibit unstable results in several cases (e.g., negative R under PURE~\cite{Stricker2014PURE}+MMPD~\cite{tang2023mmpd} or PURE~\cite{Stricker2014PURE}+UBFC-rPPG~\cite{Bobbia2017UBFCrPPG}), indicating that simple statistical alignment fails to preserve meaningful temporal dynamics.

\vspace{0.3em}
\noindent\textbf{Architecture-level transferability.} 
The consistent improvement across PhysFormer variants validates the nature of FLOW, confirming that its prototype-based temporal alignment does not depend on a specific model architecture.

\subsection{Ablation Study}
\subsubsection{Ablation on Model Components}

To investigate the contribution of each component in FLOW, we perform an ablation study by removing the Temporal Relation Module (TRM) and the Prototype-based Cross-domain Optimal Transport (PCOT) module, respectively. Experiments are conducted on BUAA-MIHR~\cite{Xi2020BUAAMIHR} and MMPD~\cite{tang2023mmpd} datasets under the same training setup as the multi-source domain generalization setting. The results are summarized in Table~\ref{tab:ablation_components}.

\begin{table}[t]
    \centering
    \caption{Ablation results on model components. "FLOW" denotes the full FLOW model with both TRM and PCOT enabled.}
    \vspace{-0.5em}
    \label{tab:ablation_components}
    \resizebox{\columnwidth}{!}{
        \begin{tabular}{l cc cc}
        \toprule
        & \multicolumn{2}{c}{\textbf{BUAA}} & \multicolumn{2}{c}{\textbf{MMPD}} \\
        \cmidrule(lr){2-3}\cmidrule(lr){4-5}
        & MAE$\downarrow$ & RMSE$\downarrow$ & MAE$\downarrow$ & RMSE$\downarrow$ \\
        \midrule
        FLOW & \textbf{2.23} & \textbf{3.36} & \textbf{7.38} & \textbf{13.12} \\
        FLOW(w/o TRM) & 3.12 & 4.61 & 8.16 & 14.10 \\
        FLOW(w/o PCOT) & 4.67 & 6.13 & 10.24 & 14.94 \\
        \bottomrule
        \end{tabular}
    }
    \vspace{-1.3em}
\end{table}

\noindent\textbf{Efficacy of TRM.} \quad  Removing the Temporal Relation Module leads to a noticeable drop in performance. Specifically, MAE increases from 2.23 to 3.12 on BUAA-MIHR~\cite{Xi2020BUAAMIHR} and from 7.38 to 8.16 on MMPD~\cite{tang2023mmpd}, indicating that the TRM effectively captures temporal dependencies and mitigates temporal noise during motion variations.

\noindent\textbf{Efficacy of PCOT.} \quad  Eliminating the PCOT module causes the largest degradation among all variants, showing an MAE of 4.67 on BUAA-MIHR~\cite{Xi2020BUAAMIHR} and 10.24 on MMPD~\cite{tang2023mmpd}. This demonstrates that cross-domain prototype alignment is crucial for learning semantically consistent and domain-invariant representations, enabling the model to generalize effectively to unseen environments.

\subsubsection{Ablation on Loss Functions}

We further examine the contribution of each loss component in FLOW, including the optimal transport loss ($L_{\text{OT}}$), the prototype alignment loss ($L_{\text{align}}$), and the semantic consistency loss ($L_{\text{sc}}$). As shown in Table~\ref{tab:ablation}, using only a single objective yields limited improvement, since each focuses on a different aspect of domain adaptation. The optimal transport term facilitates temporal correspondence between prototype distributions, while $L_{\text{align}}$ enhances cross-domain feature alignment. Adding $L_{\text{sc}}$ further preserves semantic relationships among prototypes, ensuring that inter-domain structural consistency is maintained. When all three losses are combined, the model achieves the best performance (MAE = 2.23 on BUAA-MIHR~\cite{Xi2020BUAAMIHR}, 7.38 on MMPD~\cite{tang2023mmpd}), indicating that the joint optimization effectively balances global distribution alignment and semantic-level regularization, leading to stronger domain generalization .

\begin{table}[t]
  \centering
  \caption{Ablation study on loss functions.}
  \vspace{-0.5em}
  \label{tab:ablation}
  \resizebox{\columnwidth}{!}{%
  \begin{tabular}{ccc|cccc}
    \toprule
    $L_{\text{OT}}$ & $L_{\text{id}}$ & $L_{\text{sc}}$
    & \multicolumn{2}{c}{\textbf{BUAA}}
    & \multicolumn{2}{c}{\textbf{MMPD}} \\
    \cmidrule(lr){4-5} \cmidrule(l){6-7}
    & & & MAE$\downarrow$ & RMSE$\downarrow$ & MAE$\downarrow$ & RMSE$\downarrow$ \\
    \midrule
    $\checkmark$ &             &             & 4.63 & 6.32  & 9.73 & 14.32 \\
                 & $\checkmark$ &             & 4.57 & 5.89 & 9.82 & 13.96 \\
    $\checkmark$ & $\checkmark$ &             & 2.44 & 4.12 & 8.23 & 14.42 \\
    $\checkmark$ &             & $\checkmark$ & 3.63 & 5.61 & 7.98 & 13.17 \\
                 & $\checkmark$ & $\checkmark$ & 3.77 & 5.72 & 8.87 & 14.71 \\
    $\checkmark$ & $\checkmark$ & $\checkmark$ & \textbf{2.23} & \textbf{3.36} & \textbf{7.38} & \textbf{13.12} \\
    \bottomrule
  \end{tabular}%
  }
\end{table}

\subsubsection{Ablation on Backbone}

\begin{figure}[t]
  \centering
  \includegraphics[width=0.98\linewidth]{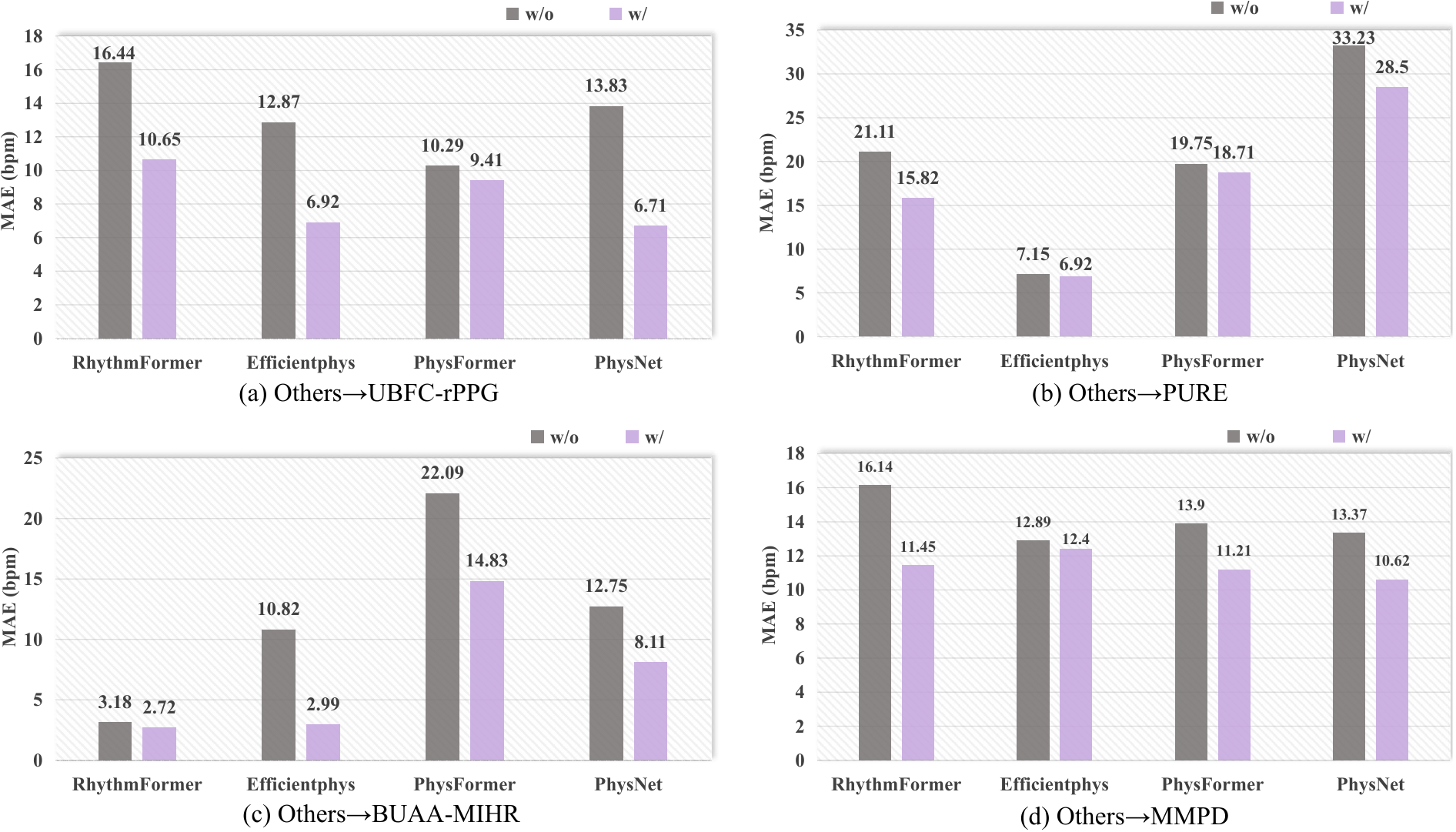}
  \vspace{-0.6em}
  \caption{Visualization of FLOW’s effectiveness across different backbones and cross-domain settings. }
  \label{fig:fig1}
  \vspace{-1.2em}
\end{figure}

To further illustrate the effectiveness and generalization capability of FLOW, we visualize its impact on different backbone models. As shown in Fig.~\ref{fig:fig1}, we compare the mean absolute error (MAE) before and after integrating FLOW into four representative architectures, RhythmFormer~\cite{zou2025rhythmformer}, EfficientPhys~\cite{liu2023efficientphys}, PhysFormer~\cite{yu2022physformer}, and PhysNet~\cite{yu2019remote}—under four cross-domain settings (Others$\rightarrow$UBFC-rPPG~\cite{Bobbia2017UBFCrPPG}, Others$\rightarrow$PURE, Others$\rightarrow$BUAA-MIHR~\cite{Xi2020BUAAMIHR}, and Others$\rightarrow$MMPD~\cite{tang2023mmpd}).

Across all backbones and datasets, incorporating FLOW consistently reduces the prediction error. Notably, PhysFormer and EfficientPhys exhibit substantial improvements on challenging domains such as BUAA-MIHR~\cite{Xi2020BUAAMIHR} and MMPD~\cite{tang2023mmpd}, where domain shifts due to illumination and motion are most severe. This confirms that FLOW effectively captures domain-invariant temporal dependencies and semantic consistency without relying on specific architectural designs. The consistent downward trend across all models highlights FLOW’s universality as a plug-and-play module for cross-domain physiological signal estimation.

\subsubsection{Analysis of Structural Hyperparameters}

\begin{figure}[t]
    \centering
    \begin{subfigure}[th]{0.47\linewidth}
        \centering
        \includegraphics[width=\linewidth]{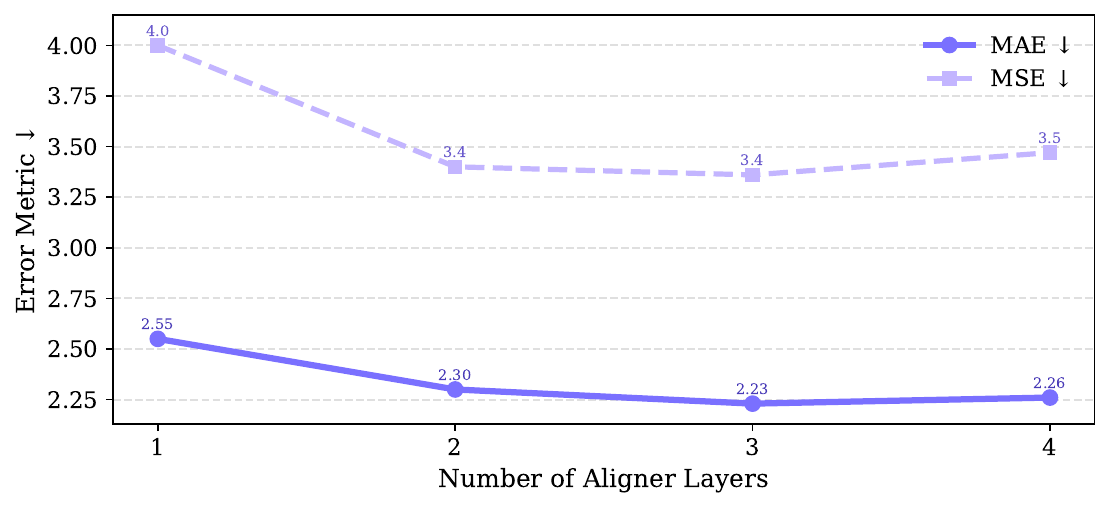}
        \vspace{-0.5em}
        \caption{Effect of the number of prototypes $K$ in the PCOT module on MAE/MSE performance.}
        \label{fig:num_prototypes_mae}
    \end{subfigure}
    \hfill
    \begin{subfigure}[t]{0.47\linewidth}
        \centering
        \includegraphics[width=\linewidth]{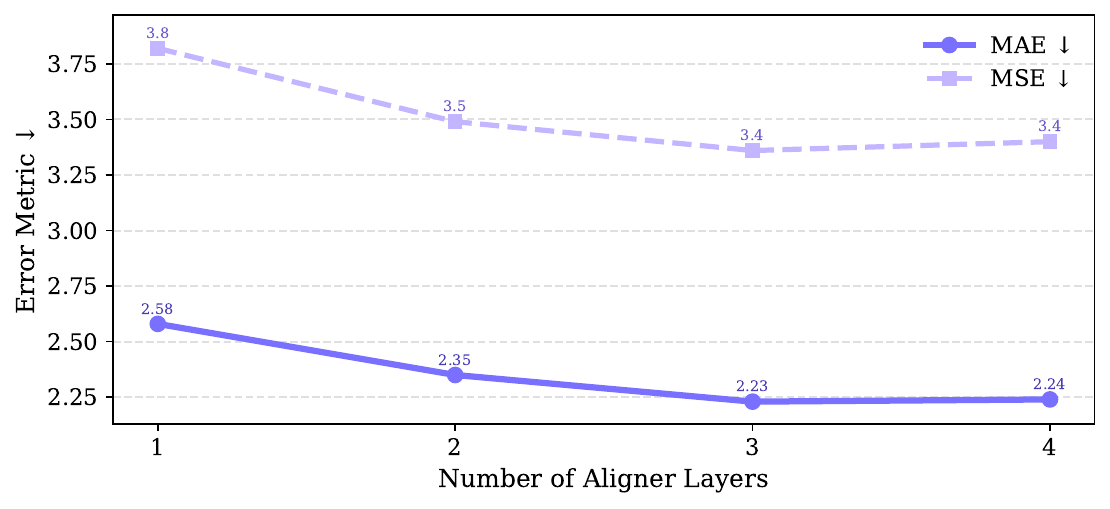}
        \vspace{-0.5em}
        \caption{Effect of the number of aligner layers in the TRM block on MAE/MSE performance.}
        \label{fig:num_aligners_mae}
    \end{subfigure}
    \vspace{-0.6em}
    \caption{Ablations on the structure of PCOT and TRM modules.}
    \label{fig:ablation_pcot_trm}
    \vspace{-1.2em}
\end{figure}

We further investigate the impact of two key structural hyperparameters in FLOW: the number of prototypes $K$ in the PCOT module and the number of aligner layers in the TRM block. As shown in Fig.~\ref{fig:num_prototypes_mae} and Fig.~\ref{fig:num_aligners_mae}, both factors influence performance but exhibit stable trends, demonstrating the robustness of our framework.

For the prototype number $K$, the MAE gradually decreases as $K$ increases from 16 to 64, reaching the optimal performance (2.23 bpm) at $K=64$. Beyond this point, the improvement saturates and slightly declines, likely due to over-fragmentation of prototype distributions that weakens semantic compactness. For the number of aligner layers, increasing layers improves representation alignment up to three layers, where the MAE reaches its minimum (2.23 bpm). Further stacking yields marginal gains, indicating that moderate depth is sufficient to capture cross-domain temporal dependencies without overfitting.

These results confirm that FLOW maintains consistent generalization across reasonable hyperparameter ranges, highlighting its stability and scalability under different architectural configurations.

\section{Conclusion}

In this paper, we propose the \textbf{FLOW}, a unified framework for domain-generalized physiological signal estimation. By combining the Temporal Refinement Module (TRM) for temporal coherence and the Prototype-based Cross-domain Optimal Transport (PCOT) for semantic alignment, FLOW effectively learns domain-invariant yet physiologically consistent representations. Extensive experiments demonstrate that FLOW achieves state-of-the-art generalization across diverse datasets and architectures. In the future, we plan to extend FLOW to multi-modal sensing and real-world deployment, advancing reliable physiological estimation under unconstrained conditions.

{
    \small
    \bibliographystyle{ieeenat_fullname}
    \bibliography{main}

@String(CVPR= {IEEE Conf. Comput. Vis. Pattern Recog.})

@String(ICCV= {Int. Conf. Comput. Vis.})

@String(ECCV= {Eur. Conf. Comput. Vis.})

@String(BMVC= {Brit. Mach. Vis. Conf.})

@String(AAAI = {AAAI})

@String(CVPR  = {CVPR})

@String(ICCV  = {ICCV})

@String(ECCV  = {ECCV})

@String(BMVC  =	{BMVC})

@inproceedings{chen2018deepphys,
  title={DeepPhys: Video-based physiological measurement using convolutional attention networks},
  author={Chen, Wei-Ting and McDuff, Daniel and Hernandez, Javier and Picard, Rosalind W},
  booktitle={Proceedings of the European Conference on Computer Vision (ECCV)},
  pages={349--365},
  year={2018}
}

@inproceedings{yu2019remote,
  title={Remote photoplethysmograph signal measurement from facial videos using spatio-temporal networks},
  author={Yu, Zhenghan and Balakrishnan, Geetha and Zhao, Xuhua and Li, Qiang and Velipasalar, Senem and Wu, Min},
  booktitle={British Machine Vision Conference (BMVC)},
  year={2019}
}

@inproceedings{niu2020video,
  title={Video-based physiological measurement via cross-verified feature disentangling},
  author={Niu, Xiaobai and Han, Hui and Shan, Shiguang and Chen, Xilin},
  booktitle={European Conference on Computer Vision (ECCV)},
  pages={545--561},
  year={2020}
}

@inproceedings{wang2021domain,
  title={Domain adaptation for remote photoplethysmography under inconsistent light conditions},
  author={Wang, Weixuan and Peng, Kunlin and Lu, Yao and Wu, Min},
  booktitle={IEEE Transactions on Instrumentation and Measurement},
  volume={70},
  pages={1--13},
  year={2021}
}

@inproceedings{yu2021physformer,
  title={PhysFormer: Facial video-based physiological measurement with temporal difference transformer},
  author={Yu, Zhenghan and Zhao, Xuhua and Balakrishnan, Geetha and et al.},
  booktitle={Proceedings of the IEEE/CVF International Conference on Computer Vision (ICCV)},
  pages={135--144},
  year={2021}
}

@article{verkruysse2008remote,
  title={Remote plethysmographic imaging using ambient light},
  author={Verkruysse, Wim and Svaasand, Lars O and Nelson, J Stuart},
  journal={Optics express},
  volume={16},
  number={26},
  pages={21434--21445},
  year={2008}
}

@inproceedings{poh2010non,
  title={Non-contact, automated cardiac pulse measurements using video imaging and blind source separation},
  author={Poh, Ming-Zher and McDuff, Daniel J and Picard, Rosalind W},
  booktitle={International Conference of the IEEE Engineering in Medicine and Biology Society (EMBC)},
  pages={3130--3133},
  year={2010}
}

@article{li2018domain,
  title={Domain generalization via conditional invariant adversarial networks},
  author={Li, Da and Yang, Yongxin and Song, Yuwei and Hospedales, Timothy M},
  journal={Advances in Neural Information Processing Systems (NeurIPS)},
  volume={31},
  year={2018}
}

@inproceedings{courty2017joint,
  title={Joint distribution optimal transportation for domain adaptation},
  author={Courty, Nicolas and Flamary, R{\'e}mi and Tuia, Devis and Rakotomamonjy, Alain},
  booktitle={Advances in Neural Information Processing Systems (NeurIPS)},
  volume={30},
  year={2017}
}

@inproceedings{damodaran2018deepjdot,
  title={DeepJDOT: Deep joint distribution optimal transport for unsupervised domain adaptation},
  author={Damodaran, Bharath Bhushan and Kellenberger, Benjamin and Flamary, R{\'e}mi and Tuia, Devis and Courty, Nicolas},
  booktitle={European Conference on Computer Vision (ECCV)},
  pages={447--463},
  year={2018}
}

@article{montesuma2021wasserstein,
  title={Wasserstein domain generalization: theoretical foundations and algorithms},
  author={Montesuma, Caio and Scardapane, Simone and Ballan, Lamberto and Dragotti, Pier Luigi},
  journal={IEEE Transactions on Pattern Analysis and Machine Intelligence (TPAMI)},
  year={2021},
  publisher={IEEE}
}

@article{mcduff2018survey,
  title={A survey of remote optical photoplethysmographic imaging methods},
  author={McDuff, Daniel},
  journal={IEEE Transactions on Biomedical Engineering},
  volume={66},
  number={4},
  pages={1--16},
  year={2018}
}

@article{zhou2021domain,
  title={Domain generalization: A survey},
  author={Zhou, Kaiyang and Yang, Yongxin and Qiao, Yu and Xiang, Tao},
  journal={IEEE Transactions on Pattern Analysis and Machine Intelligence (TPAMI)},
  year={2021},
  publisher={IEEE},
  doi={10.1109/TPAMI.2021.3059205}
}

@inproceedings{dubey2021adaptive,
  title={Adaptive risk minimization: A meta-learning approach for domain generalization},
  author={Dubey, Abhishek and Giryes, Raja and Wolf, Lior},
  booktitle={IEEE International Conference on Computer Vision (ICCV)},
  pages={1090--1099},
  year={2021}
}

@article{bobbia2019unsupervised,
  title={Unsupervised skin tissue segmentation for remote photoplethysmography},
  author={Bobbia, Serge and Macwan, Richard and Benezeth, Yannick and Mansouri, Alamin and Dubois, Julien},
  journal={Pattern recognition letters},
  volume={124},
  pages={82--90},
  year={2019},
  publisher={Elsevier}
}

@inproceedings{stricker2014non,
  title={Non-contact video-based pulse rate measurement on a mobile service robot},
  author={Stricker, Ronny and M{\"u}ller, Steffen and Gross, Horst-Michael},
  booktitle={The 23rd IEEE International Symposium on Robot and Human Interactive Communication},
  pages={1056--1062},
  year={2014},
  organization={IEEE}
}

@inproceedings{xi2020image,
  title={Image enhancement for remote photoplethysmography in a low-light environment},
  author={Xi, Lin and Chen, Weihai and Zhao, Changchen and Wu, Xingming and Wang, Jianhua},
  booktitle={2020 15th IEEE International Conference on Automatic Face and Gesture Recognition (FG 2020)},
  pages={1--7},
  year={2020},
  organization={IEEE}
}

@inproceedings{tang2023mmpd,
  title={Mmpd: Multi-domain mobile video physiology dataset},
  author={Tang, Jiankai and Chen, Kequan and Wang, Yuntao and Shi, Yuanchun and Patel, Shwetak and McDuff, Daniel and Liu, Xin},
  booktitle={2023 45th Annual International Conference of the IEEE Engineering in Medicine \& Biology Society (EMBC)},
  pages={1--5},
  year={2023},
  organization={IEEE}
}

@inproceedings{sun2022contrast,
  title={Contrast-phys: Unsupervised video-based remote physiological measurement via spatiotemporal contrast},
  author={Sun, Zhaodong and Li, Xiaobai},
  booktitle={European Conference on Computer Vision},
  pages={492--510},
  year={2022},
  organization={Springer}
}

@inproceedings{yu2019remote1,
  title={Remote photoplethysmograph signal measurement from facial videos using spatio-temporal networks},
  author={Yu, Zitong and Li, Xiaobai and Zhao, Guoying},
  booktitle={BMVC},
  year={2019}
}

@article{liu2023rppg,
  title={rppg-toolbox: Deep remote ppg toolbox},
  author={Liu, Xin and Narayanswamy, Girish and Paruchuri, Akshay and Zhang, Xiaoyu and Tang, Jiankai and Zhang, Yuzhe and Sengupta, Roni and Patel, Shwetak and Wang, Yuntao and McDuff, Daniel},
  journal={Advances in Neural Information Processing Systems},
  volume={36},
  pages={68485--68510},
  year={2023}
}

@article{peyre2019computational,
  title={Computational Optimal Transport},
  author={Peyr{\'e}, Gabriel and Cuturi, Marco},
  journal={Foundations and Trends in Machine Learning},
  volume={11},
  number={5-6},
  pages={355--607},
  year={2019},
  publisher={Now Publishers}
}

@inproceedings{cuturi2013sinkhorn,
  title={Sinkhorn distances: Lightspeed computation of optimal transport},
  author={Cuturi, Marco},
  booktitle={Advances in Neural Information Processing Systems (NeurIPS)},
  volume={26},
  pages={2292--2300},
  year={2013}
}

@article{courty2017optimal,
  title={Optimal transport for domain adaptation},
  author={Courty, Nicolas and Flamary, R{\'e}mi and Tuia, Devis and Rakotomamonjy, Alain},
  journal={IEEE Transactions on Pattern Analysis and Machine Intelligence},
  volume={39},
  number={9},
  pages={1853--1865},
  year={2017},
  publisher={IEEE}
}

@inproceedings{song2021hr,
  title={HR-CNN: Deep learning for remote heart rate estimation from face videos},
  author={Song, Jang-Han and Kim, Youngbin and Lee, Sangsoo and Lee, Seungmoon},
  booktitle={Proceedings of the IEEE/CVF Conference on Computer Vision and Pattern Recognition (CVPR)},
  pages={772--781},
  year={2021}
}

@article{xie2025physllm,
  title={PhysLLM: Harnessing Large Language Models for Cross-Modal Remote Physiological Sensing},
  author={Xie, Yiping and Zhao, Bo and Dai, Mingtong and Zhou, Jian-Ping and Sun, Yue and Tan, Tao and Xie, Weicheng and Shen, Linlin and Yu, Zitong},
  journal={arXiv preprint arXiv:2505.03621},
  year={2025}
}

@article{de2013robust,
  title={Robust pulse rate from chrominance-based rPPG},
  author={De Haan, Gerard and Jeanne, Vincent},
  journal={IEEE transactions on biomedical engineering},
  volume={60},
  number={10},
  pages={2878--2886},
  year={2013},
  publisher={IEEE}
}

@article{wang2016algorithmic,
  title={Algorithmic principles of remote PPG},
  author={Wang, Wenjin and Den Brinker, Albertus C and Stuijk, Sander and De Haan, Gerard},
  journal={IEEE Transactions on Biomedical Engineering},
  volume={64},
  number={7},
  pages={1479--1491},
  year={2016},
  publisher={IEEE}
}

@inproceedings{liu2023efficientphys,
  title={Efficientphys: Enabling simple, fast and accurate camera-based cardiac measurement},
  author={Liu, Xin and Hill, Brian and Jiang, Ziheng and Patel, Shwetak and McDuff, Daniel},
  booktitle={Proceedings of the IEEE/CVF winter conference on applications of computer vision},
  pages={5008--5017},
  year={2023}
}

@inproceedings{yu2022physformer,
  title={Physformer: Facial video-based physiological measurement with temporal difference transformer},
  author={Yu, Zitong and Shen, Yuming and Shi, Jingang and Zhao, Hengshuang and Torr, Philip HS and Zhao, Guoying},
  booktitle={Proceedings of the IEEE/CVF conference on computer vision and pattern recognition},
  pages={4186--4196},
  year={2022}
}

@article{zou2025rhythmformer,
  title={RhythmFormer: Extracting patterned rPPG signals based on periodic sparse attention},
  author={Zou, Bochao and Guo, Zizheng and Chen, Jiansheng and Zhuo, Junbao and Huang, Weiran and Ma, Huimin},
  journal={Pattern Recognition},
  volume={164},
  pages={111511},
  year={2025},
  publisher={Elsevier}
}

@article{sun2016return,
  title={Return of frustratingly easy domain adaptation},
  author={Sun, Baochen and Feng, Jiashi and Saenko, Kate},
  journal={arXiv preprint arXiv:1511.05547},
  year={2016}
}

@article{gretton2012kernel,
  title={A kernel two-sample test},
  author={Gretton, Arthur and Borgwardt, Karsten M and Rasch, Malte J and Sch{\"o}lkopf, Bernhard and Smola, Alexander},
  journal={The Journal of Machine Learning Research},
  volume={13},
  number={1},
  pages={723--773},
  year={2012}
}

@inproceedings{lu2023neuron,
  title={Neuron structure modeling for generalizable remote physiological measurement},
  author={Lu, Hao and Yu, Zitong and Niu, Xuesong and Chen, Ying-Cong},
  booktitle={Proceedings of the IEEE/CVF conference on computer vision and pattern recognition},
  pages={18589--18599},
  year={2023}
}

@inproceedings{wang2024rppg,
  title={rPPG-HiBa: Hierarchical Balanced Framework for Remote Physiological Measurement},
  author={Wang, Yin and Lu, Hao and Chen, Ying-Cong and Kuang, Li and Zhou, Mengchu and Deng, Shuiguang},
  booktitle={Proceedings of the 32nd ACM International Conference on Multimedia},
  pages={2982--2991},
  year={2024}
}

@article{wang2025physmle,
  title={Physmle: Generalizable and priors-inclusive multi-task remote physiological measurement},
  author={Wang, Jiyao and Lu, Hao and Wang, Ange and Yang, Xiao and Chen, Yingcong and He, Dengbo and Wu, Kaishun},
  journal={IEEE Transactions on Pattern Analysis and Machine Intelligence},
  year={2025},
  publisher={IEEE}
}

@inproceedings{li2024bi,
  title={Bi-TTA: Bidirectional Test-Time Adapter for Remote Physiological Measurement},
  author={Li, Haodong and Lu, Hao and Chen, Ying-Cong},
  booktitle={European Conference on Computer Vision},
  pages={356--374},
  year={2024},
  organization={Springer}
}

@article{niu2019rhythmnet,
  title={Rhythmnet: End-to-end heart rate estimation from face via spatial-temporal representation},
  author={Niu, Xuesong and Shan, Shiguang and Han, Hu and Chen, Xilin},
  journal={IEEE Transactions on Image Processing},
  volume={29},
  pages={2409--2423},
  year={2019},
  publisher={IEEE}
}

@article{gangbo1996geometry,
  title={The geometry of optimal transportation},
  author={Gangbo, Wilfrid and McCann, Robert J},
  year={1996}
}

@article{Bobbia2017UBFCrPPG,
  title        = {Unsupervised skin tissue segmentation for remote photoplethysmography},
  author       = {Bobbia, Serge and Macwan, Richard and Benezeth, Yannick and Mansouri, Alamin and Dubois, Julien},
  journal      = {Pattern Recognition Letters},
  volume       = {124},
  pages        = {82--90},
  year         = {2017},
  doi          = {10.1016/j.patrec.2017.10.017}
}

@inproceedings{Stricker2014PURE,
  author    = {Stricker, Ronny and Mueller, Steffen and Gross, Horst{-}Michael},
  title     = {Non-contact Video-based Pulse Rate Measurement on a Mobile Service Robot},
  booktitle = {IEEE Int. Symposium on Robot and Human Interactive Communication (RO-MAN)},
  pages     = {1056--1062},
  year      = {2014},
  publisher = {IEEE}
}

@inproceedings{Xi2020BUAAMIHR,
  author    = {Xi, Lin and Chen, Weihai and Zhao, Changchen and Wu, Xingming and Wang, Jianhua},
  title     = {Image Enhancement for Remote Photoplethysmography in a Low-Light Environment},
  booktitle = {2020 15th IEEE International Conference on Automatic Face and Gesture Recognition (FG 2020)},
  pages     = {1--7},
  year      = {2020},
  publisher = {IEEE}
}

@article{zhang2025advancing,
  title={Advancing generalizable remote physiological measurement through the integration of explicit and implicit prior knowledge},
  author={Zhang, Yuting and Lu, Hao and Liu, Xin and Chen, Yingcong and Wu, Kaishun},
  journal={IEEE Transactions on Image Processing},
  year={2025},
  publisher={IEEE}
}

@InProceedings{Du_2023_CVPR,
  author    = {Du, Jingda and Liu, Si-Qi and Zhang, Bochao and Yuen, Pong C.},
  title     = {Dual-Bridging With Adversarial Noise Generation for Domain Adaptive rPPG Estimation},
  booktitle = {Proceedings of the IEEE/CVF Conference on Computer Vision and Pattern Recognition (CVPR)},
  month     = {June},
  year      = {2023},
  pages     = {10355–10364}
}

@InProceedings{Savic_Zhao_2024_ECCV,
  author    = {Savic, Marko and Zhao, Guoying},
  title     = {Oulu Remote-Photoplethysmography Physical Domain Attacks Database (ORPDAD)},
  booktitle = {Computer Vision – ECCV 2024, Lecture Notes in Computer Science, Vol. 15131},
  editor    = {Leonardis, A. and Ricci, E. and Roth, S. and Russakovsky, O. and Sattler, T. and Varol, G.},
  publisher = {Springer, Cham},
  year      = {2024},
  pages     = {51–68},
  doi       = {10.1007/978-3-031-73464-9_4}
}

@Article{Xie_et_al_2024_arXiv,
  author    = {Xie, Yiping and Yu, Zitong and Wu, Bingjie and Xie, Weicheng and Shen, Linlin},
  title     = {SFDA-rPPG: Source-Free Domain Adaptive Remote Physiological Measurement with Spatio-Temporal Consistency},
  journal   = {arXiv preprint},
  year      = {2024},
  note      = {arXiv:2409.12040}
}

@inproceedings{huang2024etag,
  title={etag: Class-incremental learning via embedding distillation and task-oriented generation},
  author={Huang, Libo and Zeng, Yan and Yang, Chuanguang and An, Zhulin and Diao, Boyu and Xu, Yongjun},
  booktitle={Proceedings of the AAAI Conference on Artificial Intelligence},
  volume={38},
  number={11},
  pages={12591--12599},
  year={2024}
}

@article{huang2021unified,
  title={A unified optimization model of feature extraction and clustering for spike sorting},
  author={Huang, Libo and Gan, Lu and Ling, Bingo Wing-Kuen},
  journal={IEEE Transactions on Neural Systems and Rehabilitation Engineering},
  volume={29},
  pages={750--759},
  year={2021},
  publisher={IEEE}
}

@article{zeng2024survey,
  title={A survey on causal reinforcement learning},
  author={Zeng, Yan and Cai, Ruichu and Sun, Fuchun and Huang, Libo and Hao, Zhifeng},
  journal={IEEE Transactions on Neural Networks and Learning Systems},
  year={2024},
  publisher={IEEE}
}

@article{liu2024one,
  title={One-shot Face Reenactment with Dense Correspondence Estimation},
  author={Liu, Yunfan and Li, Qi and Sun, Zhenan},
  journal={Machine Intelligence Research},
  volume={21},
  number={5},
  pages={941--953},
  year={2024},
  publisher={Springer}
}

@article{shi2024adaptively,
  title={Adaptively enhancing facial expression crucial regions via a local non-local joint network},
  author={Shi, Guanghui and Mao, Shasha and Gou, Shuiping and Yan, Dandan and Jiao, Licheng and Xiong, Lin},
  journal={Machine Intelligence Research},
  volume={21},
  number={2},
  pages={331--348},
  year={2024},
  publisher={Springer}
}

@inproceedings{liu2025llm,
  title={Au-llm: Micro-expression action unit detection via enhanced llm-based feature fusion},
  author={Liu, Zhishu and Yuan, Kaishen and Zhao, Bo and Xu, Yong and Yu, Zitong},
  booktitle={Chinese Conference on Biometric Recognition},
  pages={355--365},
  year={2025},
  organization={Springer}
}

@article{liu2026aullm++,
  title={AULLM++: Structural Reasoning with Large Language Models for Micro-Expression Recognition},
  author={Liu, Zhishu and Yuan, Kaishen and Zhao, Bo and Ma, Hui and Yu, Zitong},
  journal={arXiv preprint arXiv:2603.08387},
  year={2026}
}

@inproceedings{zhu2026H-GAR,
  title={H-GAR: A Hierarchical Interaction Framework via Goal-Driven Observation-Action Refinement for Robotic Manipulation},
  author={Zhu, Yijie and Shao, Rui and Liu, Ziyang and He, Jie and Liu, Jizhihui and Wang, Jiuru and Yu, Zitong},
  booktitle={Proceedings of the AAAI Conference on Artificial Intelligence},
  year={2026}
}

@inproceedings{zhu2025emosym,
  title={EmoSym: A Symbiotic Framework for Unified Emotional Understanding and Generation via Latent Reasoning},
   author={Zhu, Yijie and Lyu, Yibo and Yu, Zitong and Shao, Rui and Zhou, Kaiyang and Nie, Liqiang},
  booktitle={Proceedings of the 33nd ACM International Conference on Multimedia},
  year={2025}
}

@article{zhu2025uniemo,
  title={UniEmo: Unifying Emotional Understanding and Generation with Learnable Expert Queries},
  author={Zhu, Yijie and Zhang, Lingsen and Yu, Zitong and Shao, Rui and Tan, Tao and Nie, Liqiang},
  journal={arXiv preprint arXiv:2507.23372},
  year={2025}
}

@article{cao2026physnext,
  title={PhysNeXt: Next-Generation Dual-Branch Structured Attention Fusion Network for Remote Photoplethysmography Measurement},
  author={Cao, Junzhe and Zhao, Bo and Niu, Zhiyi and Guo, Dan and Sun, Yue and Liang, Haochen and Xu, Yong and Yu, Zitong},
  journal={arXiv preprint arXiv:2603.19752},
  year={2026}
}

@article{huang2026complementarity,
  title={Complementarity-Supervised Spectral-Band Routing for Multimodal Emotion Recognition},
  author={Huang, Zhexian and Zhao, Bo and Ma, Hui and Liu, Zhishu and Zhang, Jie and Zhang, Ruixin and Ding, Shouhong and Yu, Zitong},
  journal={arXiv preprint arXiv:2603.13340},
  year={2026}
}

@article{wu2025cardiacmamba,
  title={Cardiacmamba: A multimodal rgb-rf fusion framework with state space models for remote physiological measurement},
  author={Wu, Zheng and Xie, Yiping and Zhao, Bo and He, Jiguang and Luo, Fei and Deng, Ning and Yu, Zitong},
  journal={arXiv preprint arXiv:2502.13624},
  year={2025}
}
}

\clearpage
\setcounter{page}{1}
\maketitlesupplementary

\appendix
\section{Introduction to the Datasets}
\label{sec:datasets}
\paragraph{UBFC-rPPG~\cite{Bobbia2017UBFCrPPG}} contains 42 RGB facial videos from 42 distinct subjects. Each video is captured at 640×480 pixel resolution and 30 frames per second (fps). Recordings take place under varied lighting conditions, including natural sunlight and indoor artificial illumination. Ground-truth physiological signals are recorded via a CMS50E pulse oximeter at 60 Hz, ensuring precise temporal alignment for evaluation.
\vspace{-1.5em}
\paragraph{PURE~\cite{Stricker2014PURE}} comprises 60 high-quality RGB videos collected from 10 subjects performing six different head movement scenarios (static, talking, translation movements, etc.). Videos are recorded at 30 fps under consistent indoor lighting and controlled background settings, minimizing external interference. Synchronized physiological measurements are obtained using a CMS50E oximeter sampling at 60 Hz. PURE is particularly valuable for evaluating rPPG performance during facial movements.
\vspace{-1.5em}
\paragraph{BUAA-MIHR~\cite{Xi2020BUAAMIHR}}  is designed to assess algorithmic robustness across varying illumination intensities. The dataset features video sequences recorded under a range of controlled lighting conditions, from low-light (below 10 lux) to normal brightness. In our experiments, we only utilize videos captured under illumination levels $\geq$10 lux, as extremely dim lighting introduces significant image degradation requiring specialized enhancement techniques beyond this study's scope.
\vspace{-1.5em}
\paragraph{MMPD~\cite{tang2023mmpd}} comprises 660 videos, each lasting one minute, collected from 33 subjects with diverse skin tones and gender distributions. Each video is recorded at 30 fps with a resolution of 320×240 pixels, under four distinct lighting conditions (bright, warm, dim, and colored lighting). Subjects perform various daily activities, introducing intra-subject variability and further increasing dataset complexity.

\section{Experimental Settings}
\label{settings}
\paragraph{Datasets and Evaluation Metrics}
We evaluate our method on four widely used remote photoplethysmography (rPPG) datasets: \textbf{UBFC-rPPG}~\cite{bobbia2019unsupervised},  \textbf{PURE}~\cite{stricker2014non},  \textbf{BUAA-MIHR}~\cite{xi2020image} and \textbf{MMPD}~\cite{tang2023mmpd}.
Following prior works~\cite{sun2022contrast, yu2019remote1}, we adopt three standard evaluation metrics: mean absolute error (MAE), root mean square error (RMSE), and Pearson’s correlation coefficient (R), to assess the accuracy of predicted heart rates (HRs). For both MAE and RMSE, lower values indicate smaller prediction errors, while higher values of R (closer to 1.0) indicate stronger linear correlation with the ground-truth HRs. MAE and RMSE are reported in beats per minute (bpm); for brevity, we omit these units in subsequent tables and discussions.

\vspace{-1.5em}
\paragraph{Implementation Details}
Our experiments are implemented in PyTorch, primarily based on the rPPG-Toolbox~\cite{liu2023rppg}. For preprocessing, we detect and crop the face region from the first frame of each video clip and apply a fixed bounding box across subsequent frames. Each video is resampled to a consistent frame rate of 30 fps, and a random chunk of 128 frames is selected, resized to $128 \times 128$ pixels. We adopt the PhysLLM~\cite{xie2025physllm} as the baseline. The hyperparameters $\alpha = 0.8$ and $l_{target} = 32$ are set by default. The LLM is trained using the Adam optimizer with an initial learning rate of $1 \times 10^{-4}$ and a weight decay of $5 \times 10^{-5}$. The entire model is trained for 20 epochs on an NVIDIA H100 GPU with a batch size of 4.

\section{Theoretical Proof of the Generalization Bound}
\label{a}
\noindent
In this section, we provide a detailed derivation of the proposed
multi-source generalization bound under the conditional optimal
transport (OT) geometry used in FLOW. Our goal is to connect
the quality of cross-domain alignment---measured under a
task-aware OT cost---to the prediction risk on an unseen
target domain.

\subsection*{Preliminaries}

\noindent
\textbf{Data and hypothesis space.}
Let $\{D_s\}_{s=1}^m$ be the source domains with
mixing coefficients $\alpha_s \ge 0$ such that
$\sum_{s=1}^m \alpha_s = 1$, and let $T$ denote the
(target) test domain.
Each sample is denoted by
$z = (x, hr)$, where $x \in \mathbb{R}^d$ is the
visual-temporal feature and $hr \in \mathbb{R}$ is the
continuous heart rate label.
A hypothesis $h \in \mathcal{H}$ produces a prediction
$\hat{y}$ from $x$. We assume a bounded loss
$\ell(h(x), y) \in [0, 1]$.
For a distribution $P$, the expected risk is
\begin{equation}
    R_P(h) = \mathbb{E}_{(z,y)\sim P}[\ell(h(x), y)].
\end{equation}

\noindent
\textbf{Task-driven conditional cost.}
Following our formulation in the main paper, we define a
conditional ground cost $c$ that jointly accounts for feature
distance and physiological consistency:
\begin{equation}
    c\big((x, hr), (x', hr')\big)
    =
    \|x - x'\|_W^2
    +
    \lambda_{hr}
    \Big(
        1 - \exp\!\big(-\tfrac{(hr - hr')^2}{2\sigma^2}\big)
    \Big),
\end{equation}
where $W = \mathrm{diag}(w_1,\dots,w_d)$ is the learnable
frequency/feature weighting matrix, and
$\lambda_{hr}, \sigma > 0$ control the influence and
bandwidth of the heart-rate kernel term.
This cost emphasizes both semantic similarity in feature space
and coherence in underlying physiological signals.

\noindent
\textbf{Conditional OT distance.}
Given two distributions $P,Q$ on $(x, hr)$, we define the
conditional OT distance:
\begin{equation}
    W_c(P,Q)
    =
    \inf_{\pi \in \Pi(P,Q)}
    \int c(z,z') \, d\pi(z,z'),
\end{equation}
where $\Pi(P,Q)$ is the set of couplings with marginals
$P$ and $Q$. Throughout, we assume $c$ induces a valid
(or pseudo-)metric structure compatible with the OT geometry.

\noindent
\textbf{OT barycenter over sources.}
We consider the (conditional) OT barycenter $B$ of
$\{D_s\}_{s=1}^m$:
\begin{equation}
    B
    =
    \arg\min_{\nu}
    \sum_{s=1}^m \alpha_s W_c(D_s, \nu).
\end{equation}
Intuitively, $B$ captures a geometry-aware “anchor”
distribution that balances all source domains under $W_c$.

\medskip
\noindent
\textbf{Regularized OT and residual terms.}
In practice, we employ a debiased Sinkhorn divergence
to approximate $W_c$, together with structure-preserving
regularization terms (e.g., identity-preserving constraints)
used in FLOW. We collect these deviations into:
\begin{itemize}
    \item $\Delta_{\text{sink}}$:
    the residual bias between the ideal $W_c$ and its
    debiased Sinkhorn approximation;
    \item $\Delta_{\text{id}}$:
    the residual bias introduced by identity-preserving
    regularization (and related structure-preserving penalties)
    that slightly perturb the ideal OT geometry.
\end{itemize}
Both $\Delta_{\text{sink}}$ and $\Delta_{\text{id}}$ are
treated as small non-negative constants controlled by
regularization strength and optimization accuracy.

\subsection*{Risk Discrepancy under Conditional OT}

\noindent
We first relate the risk difference between two domains to their
conditional OT distance.

\medskip
\noindent
\textbf{Lemma 1 (Task-related discrepancy bound).}
Let
$g(z) = \mathbb{E}[\ell(h(x), y) \mid z]$.
Assume $g$ is $L_c$-Lipschitz with respect to the metric
induced by $c$, i.e.,
\begin{equation}
    |g(z) - g(z')|
    \le
    L_c \, d_c(z,z')
    \le
    L_c \, c(z,z').
\end{equation}
Then for any distributions $P,Q$,
\begin{equation}
    |R_P(h) - R_Q(h)|
    \le
    L_c \, W_c(P,Q) + \Delta_{\text{id}}.
\end{equation}

\noindent
\textit{Proof.}
Let $\pi^\star \in \Pi(P,Q)$ be an optimal coupling for $W_c$.
Then
\begin{align}
    R_P(h) - R_Q(h)
    &=
    \mathbb{E}_{P}[g(z)] - \mathbb{E}_{Q}[g(z')]
    =
    \iint (g(z) - g(z')) \, d\pi^\star(z,z').
\end{align}
By the Lipschitz property,
$|g(z) - g(z')| \le L_c \, c(z,z')$,
thus
\begin{align}
    |R_P(h) - R_Q(h)|
    &\le
    L_c \iint c(z,z') \, d\pi^\star(z,z')
    =
    L_c W_c(P,Q)
    + \Delta_{\text{id}},
\end{align}
where $\Delta_{\text{id}}$ accounts for the slight distortion
introduced by identity-preserving regularization in the
practical alignment.
\hfill $\square$

\subsection*{From Sources to Barycenter and Target}

\noindent
Using Lemma 1, we control the risk at the barycenter $B$
and then transfer it to the target domain $T$.

\medskip
\noindent
\textbf{Lemma 2 (Multi-source to barycenter).}
For any $h \in \mathcal{H}$,
\begin{equation}
    R_B(h)
    \le
    \sum_{s=1}^m \alpha_s R_{D_s}(h)
    +
    L_c
    \sum_{s=1}^m \alpha_s W_c(D_s, B)
    +
    \Delta_{\text{id}}.
\end{equation}

\noindent
\textit{Proof.}
Apply Lemma 1 with $(P,Q) = (D_s, B)$ and then average
with weights $\alpha_s$:
\begin{equation}
    R_B(h)
    \le
    \sum_s \alpha_s R_{D_s}(h)
    +
    L_c \sum_s \alpha_s W_c(D_s,B)
    +
    \Delta_{\text{id}}.
\end{equation}
\hfill $\square$

\medskip
\noindent
\textbf{Lemma 3 (Barycenter to target).}
For any $h \in \mathcal{H}$,
\begin{equation}
    R_T(h)
    \le
    R_B(h) + L_c W_c(B,T) + \Delta_{\text{id}}.
\end{equation}

\noindent
\textit{Proof.}
Apply Lemma 1 with $(P,Q) = (B,T)$ directly.
\hfill $\square$

\medskip
\noindent
\textbf{Lemma 4 (Two-hop barycenter inequality).}
Assume $W_c$ satisfies the triangle inequality and
$B$ is the OT barycenter defined above.
Then
\begin{equation}
    \sum_{s=1}^m \alpha_s W_c(D_s, B)
    +
    W_c(B, T)
    \le
    2
    \sum_{s=1}^m \alpha_s W_c(D_s, T).
\end{equation}

\noindent
\textit{Sketch.}
By barycenter optimality,
for any $\nu$ (in particular $\nu = T$),
\begin{equation}
    \sum_s \alpha_s W_c(D_s, B)
    \le
    \sum_s \alpha_s W_c(D_s, T).
\end{equation}
By triangle inequality,
$W_c(D_s, T) \le W_c(D_s, B) + W_c(B, T)$,
hence
\begin{equation}
    W_c(B, T)
    \le
    W_c(D_s, T) - W_c(D_s, B)
    + \text{(non-negative term)}.
\end{equation}
Aggregating over $s$ and combining the two relations yields
an upper bound of the two-hop quantity by a constant factor
of the direct discrepancies $\{W_c(D_s,T)\}$; the above
inequality is a convenient sufficient form.
A fully rigorous derivation can be obtained by exploiting
convexity of OT distances and the characterization of Wasserstein
barycenters; we omit the routine details for brevity.
\hfill $\square$

\subsection*{Conditional OT Generalization Bound}

\noindent
We now combine the above lemmas to obtain the main bound.

\medskip
\noindent
\textbf{Theorem 1 (Conditional OT barycenter generalization bound).}
For any $h \in \mathcal{H}$,
\begin{align}
    R_T(h)
    &\le
    \sum_{s=1}^m \alpha_s R_{D_s}(h)
    +
    L_c
    \Big(
        \sum_{s=1}^m \alpha_s W_c(D_s, B)
        +
        W_c(B, T)
    \Big)
    \notag\\
    &\quad
    +
    \Delta_{\text{sink}}
    +
    \Delta_{\text{id}}
    +
    \Delta_{\text{est}}.
\end{align}

where $\Delta_{\text{est}}$ denotes standard statistical
estimation errors from finite samples.

\noindent
\textit{Proof.}
Starting from Lemma 3,
\begin{equation}
    R_T(h)
    \le
    R_B(h) + L_c W_c(B, T) + \Delta_{\text{id}},
\end{equation}
then substitute Lemma 2 for $R_B(h)$ to obtain
\begin{align}
    R_T(h)
    &\le
    \sum_s \alpha_s R_{D_s}(h)
    +
    L_c \sum_s \alpha_s W_c(D_s, B)
    \notag\\
    &\quad
    +
    L_c W_c(B, T)
    +
    \Delta_{\text{id}}.
\end{align}

Approximating $W_c$ with the debiased Sinkhorn divergence
contributes an additional $\Delta_{\text{sink}}$,
and empirical estimation of risks contributes $\Delta_{\text{est}}$,
leading to the stated inequality.
\hfill $\square$

\subsection*{Comparison with Traditional Geometric Bounds}

\noindent
We compare our conditional OT-based discrepancy with the
standard Euclidean Wasserstein-1 bound for multi-source
domain generalization.

\medskip
\noindent
\textbf{Classical bound.}
A typical geometric bound using the Wasserstein-1 distance
$W_1$ has the form:
\begin{equation}
    R_T(h)
    \le
    \sum_{s=1}^m \alpha_s R_{D_s}(h)
    +
    L_1
    \sum_{s=1}^m \alpha_s W_1(D_s, T)
    +
    \lambda^\star + \tilde{\Delta},
\end{equation}
where $L_1$ is a Lipschitz constant under the Euclidean
metric, $\lambda^\star$ denotes an irreducible joint error term,
and $\tilde{\Delta}$ collects statistical/approximation errors.

\medskip
\noindent
\textbf{Theorem 2 (Relative tightness under conditional geometry).}
Assume:
(i) bounded feature domain $\|x\|\le M_x$;
(ii) bounded weight matrix $\|W\|_{\mathrm{op}} \le \Lambda_W$;
(iii) heart-rate kernel term is $L_{hr}$-Lipschitz; and
(iv) the conditional metric removes irrelevant (non-physiological)
directions so that there exists $0 < r < 1$ with
$L_c \le r L_1$.
Then there exists a constant $\kappa > 0$ such that
\begin{equation}
    W_c(P,Q) \le \kappa W_1(P,Q),
\end{equation}
and, combining Lemma~4 with Theorem~1,
our discrepancy term satisfies
\begin{align}
    \text{Diff}_{\text{ours}}
    &\;\;\lesssim\;\;
    2 L_c \sum_{s=1}^m \alpha_s W_c(D_s, T)
    \notag\\
    &\;\;\le\;\;
    2 r \kappa L_1 \sum_{s=1}^m \alpha_s W_1(D_s, T).
\end{align}

up to $(\Delta_{\text{sink}} + \Delta_{\text{id}} + \Delta_{\text{est}})$.
Under standard boundedness and normalization assumptions, both $r$ and $\kappa$ remain moderate constants. 
In particular, when $2r\kappa < 1$, which is a mild condition that can be encouraged in practice via feature normalization and regularization, the conditional OT geometry yields a strictly tighter or comparable discrepancy term than the Euclidean Wasserstein-1 counterpart.
\hfill $\square$

\section{Training and Inference Procedure of FLOW}
\label{train}
In this section, we provide additional details regarding the training and inference pipeline of \textbf{FLOW}. During training, TRM first stabilizes intermediate representations, after which PCOT computes soft cross-temporal correspondences to align features across domains, as summarized in Algorithm~\ref{alg:train}. The regularization terms jointly ensure consistent prototype usage and preserve feature identity throughout the alignment process. At inference time, FLOW operates without requiring domain labels or additional preprocessing, producing temporally coherent and domain-invariant representations that support robust physiological signal prediction, as illustrated in Algorithm~\ref{alg:inference}. For clarity, we summarize the full procedure in the following subsections.

\begin{algorithm}[!h]
\caption{Training of FLOW}
\label{alg:train}
\KwIn{Multi-source videos $\{X_i\}$, domains $\{d_i\}$, rPPG signals $\{s_i\}$;
hyper-parameters $\lambda_{OT},\lambda_{sc},\lambda_{\text{id}},\alpha,\beta,\eta,E,B$.}
\KwOut{Parameters $\theta_b,\theta_t,\theta_p,\theta_r$; prototype bank $P=\{P_d\}$.}

Initialize $f_{\text{backbone}}, f_{\text{TRM}}, \text{PCOT}, h_{\text{reg}}$, optimizer, $P$\;

\For{$epoch=1$ \KwTo $E$}{
  sample minibatch $\{(X_i,d_i,s_i)\}_{i=1}^B$\;

  \tcp{1. HR labels from FFT}
  $y_i \gets \text{FFT\_HR}(s_i)$ for all $i$\;

  \tcp{2. Backbone + TRM}
  $F_i \gets f_{\text{backbone}}(X_i;\theta_b)$\;
  $C_i \gets f_{\text{TRM}}(F_i;\theta_t)$\;
  flatten $\{C_i\}$ to $F_{\text{flat}}=\{F_n\}_{n=1}^{N}$,
  repeat $d_i,y_i$ to $d_{\text{flat}},y_{\text{flat}}$\;

  \tcp{3. PCOT with label-aware cost}
  \For{$n=1$ \KwTo $N$}{
    select domain prototypes $P_n$ from $P_{d_{\text{flat}}[n]}$\;
    \For{$j=1$ \KwTo $K$}{
      $C[n,j] = \alpha\,\mathrm{dist}(F_n,P_n[j])
              + \beta\,|y_{\text{flat}}[n]-\bar y_{P_n[j]}|$\;
    }
  }
  compute transport plan $\Pi^\star = \text{OT\_Solver}(C)$\;
  $\mathcal{L}_{OT} = \sum_{n,j}\Pi^\star_{n,j} C[n,j]$\;
  $\mathcal{L}_{sc} = \frac{1}{N}\sum_{n} \min_j \mathrm{dist}(F_n,P_n[j])^2$\;
  $A_n = \sum_j \Pi^\star_{n,j} P_n[j]$ for all $n$\;

  \tcp{4. Regression loss}
  aggregate $\{A_n\}$ by sample to get $A_i$\;
  $\hat y_i = h_{\text{reg}}(A_i;\theta_r)$\;
  $\mathcal{L}_{\text{task}} = \frac{1}{B}\sum_i (\hat y_i - y_i)^2$\;

  \tcp{5. Identity-preserving loss}
  $\mathcal{L}_{\text{id}} = \frac{1}{N}\sum_{n=1}^{N} \lVert A_n - F_n \rVert_2^2$\;

  \tcp{6. Total loss and update}
  $\mathcal{L} = \mathcal{L}_{\text{task}}
  + \lambda_{OT}\mathcal{L}_{OT}
  + \lambda_{sc}\mathcal{L}_{sc}
  + \lambda_{\text{id}}\mathcal{L}_{\text{id}}$\;
  update $\theta_b,\theta_t,\theta_p,\theta_r$ by SGD on $\mathcal{L}$\;
  $P \gets \text{UpdatePrototypes}(P,\{A_n\},d_{\text{flat}})$\;
}

\Return $\theta_b,\theta_t,\theta_p,\theta_r,P$\;
\end{algorithm}

\begin{algorithm}[!h]
\caption{Inference of FLOW}
\label{alg:inference}
\KwIn{Trained $\theta_b,\theta_t,\theta_p,\theta_r$, prototype bank $P$, test video $X$.}
\KwOut{Predicted heart rate $\hat y$.}

\tcp{1. Backbone + TRM}
$F \gets f_{\text{backbone}}(X;\theta_b)$\;
$C \gets f_{\text{TRM}}(F;\theta_t)$\;
flatten $C$ to $F_{\text{flat}}=\{F_n\}_{n=1}^{N}$\;

\tcp{2. PCOT }
\For{$n=1$ \KwTo $N$}{
  choose prototypes $P_n$ (e.g., corresponding domain or all domains)\;
  \For{$j=1$ \KwTo $K$}{
    $C[n,j] = \mathrm{dist}(F_n,P_n[j])$\;
  }
}
$\Pi = \text{OT\_Solver}(C)$ (or nearest-prototype)\;
$A_n = \sum_j \Pi_{n,j} P_n[j]$ for all $n$\;

\tcp{3. Regression}
aggregate $\{A_n\}$ to video-level $A$\;
$\hat y = h_{\text{reg}}(A;\theta_r)$\;
\Return $\hat y$\;

\end{algorithm}

\end{document}